\documentclass{article} 
\usepackage{iclr2021_conference,times}

\usepackage{amsmath,amsfonts,bm}

\def\eqref#1{equation~\ref{#1}}

\def\1{\bm{1}}

\DeclareMathAlphabet{\mathsfit}{\encodingdefault}{\sfdefault}{m}{sl}
\SetMathAlphabet{\mathsfit}{bold}{\encodingdefault}{\sfdefault}{bx}{n}

\usepackage{hyperref}
\usepackage{url}
\usepackage{algorithm}
\usepackage{subcaption}
\usepackage{graphicx}  
\usepackage[noend]{algpseudocode}

\makeatletter
\def\BState{\State\hskip-\ALG@thistlm}
\makeatother

\title{AutoGRAMS: Autonomous Graphical Agent Modeling Software}

\author{Ben Krause, Lucia Chen, Emmanuel Kahembwe \\
Autograms AI, USA \\
\texttt{{bkrause}@autograms.ai}}

\newcommand{\comm}[1]{}

\iclrfinalcopy 
\begin{document}

\maketitle
\begin{abstract}

We introduce the AutoGRAMS framework for programming multi-step interactions with language models. AutoGRAMS represents AI agents as a graph, where each node can execute either a language modeling instruction or traditional code. Likewise, transitions in the graph can be governed by either language modeling decisions or traditional branch logic. AutoGRAMS supports using variables as memory and allows nodes to call other AutoGRAMS graphs as functions. We show how AutoGRAMS can be used to design highly sophisticated agents, including self-referential agents that can modify their own graph. AutoGRAMS's graph-centric approach aids interpretability, controllability, and safety during the design, development, and deployment of AI agents. We provide our framework as open source at \url{https://github.com/autograms/autograms}.
\end{abstract}

\tableofcontents
\newpage

\section{Introduction}

Artificial intelligence (AI) agents driven by large language models (LLMs) can tackle complex, multi-faceted tasks. LLMs pretrained on large datasets generate human-like text and display knowledge of various domains \cite{radford2019language, brown2020language}. This enables them to perform a range of tasks and learn from training examples given in their prompt.

Training these language models to follow instructions enhances their utility, allowing them to generate outputs that align with given instructions \cite{ouyang2022training}. This instruction-following behavior is crucial for the development of AI agents that provide useful and contextually relevant responses across a variety of applications \citep{Yao2022,Paranjape2023,Khattab2022}. It also enables these agents to drive autonomous operations, such as making decisions, accessing external information, and engaging in conversations. Despite their potential, current approaches often struggle with maintaining coherent behavior over extended interactions, adapting to unexpected user inputs, and providing designers with intuitive and flexible control over the agent’s decision-making process.

To overcome these limitations, we introduce the \textbf{AutoGRAMS (Autonomous Graphical Agent Modeling Software)} framework to empower designers to orchestrate intricate interactions with LLMs. AutoGRAMS enables the creation of sophisticated AI agents and chatbots by representing their behavior as a series of interconnected nodes, each with distinct actions and transition rules. This graphical representation provides a programming language for agent development, where the execution path dynamically adapts based on language model predictions or predefined conditions.

With AutoGRAMS, designers gain the ability to:
\begin{itemize}
    
    \item \textbf{Design branched multi-step interactions with a language model}: Define a series of interconnected nodes that represent the steps involved in language model generation. This graph outlines a pre-defined procedure, and its nodes may include multiple branches and decision points. Interaction is defined by graph traversal, by iteratively stepping through the nodes of the graph.
    \item \textbf{Design complex conversational flows}: Define a graph of conversational reply steps with prompts that govern how the agent gives conversational replies and makes decisions about which conversational branches to take.
    \item \textbf{Control interaction steps with with node-specific prompts:} Craft prompts tailored to each node, guiding the language model's responses.
    \item \textbf{Leverage language model predictions in transitions}: Define how language model predictions are used to navigate branch points and guide the agent's behavior.

    \item \textbf{Incorporate conditional logic:} Define specific conditions under which different actions or transitions occur.
    \item \textbf{Use variables to control memory:} Define how variables in memory are set and used by a language model
    \item \textbf{Integrate code:} Define Python statements to manipulate variables or interact with external APIs at any node of the interaction
    \item \textbf{Call agent modules as functions:} Define callable agent systems that can be called at points in the interaction, allowing an interaction to return to a previous point, and giving additional control over prompt scopes.
    \item \textbf{Visualize agent behavior:} Gain a clear understanding of the agent's decision-making process and current state through an intuitive graphical interface.
    \item \textbf{Design agent interactions in spreadsheets:} Enable the design of complex agents using spreadsheet software such as Microsoft Excel, Google Sheets, etc.
\end{itemize}

The designer does not necessarily need to be a person--the high level of flexibility of AutoGRAMS makes it possible for an AutoGRAMS agent to design other AutoGRAMS agents. AutoGRAMS also contains functionality to allow agents to modify their own graph directly, opening up possibilities for agents that learn by modifying their own behavior.

By bridging the gap between intuitive design and complex agent behavior, AutoGRAMS opens up new possibilities for developing AI agents that can effectively navigate diverse scenarios or tasks, adapt to user input, and maintain coherent, goal-oriented conversations. Specifically, by enabling designers to define the behavior of individual nodes and transitions flexibly and easily within a graphical framework, AutoGRAMS facilitates precise control over multi-step interactions with language models. This versatile tool serves as both a high-level programming language and a comprehensive framework for developing language model-based agents and chatbots.

The layout of this paper is as follows: We cover the AutoGRAMS framework in Section \ref{sec:framework}. We demonstrate the mapping from python-like code and AutoGRAMS, along with the AutoGRAMS compiler--in Section \ref{sec:compiler}. We introduce how AutoGRAMS agents are executed in Section \ref{sec:interpreter}. And finally, we discuss self-modifying agents and agents that can design agents in Section \ref{sec:meta-self}.

\section{AutoGRAMS Framework}
\label{sec:framework}

 We introduce AutoGRAMS and go over the process of defining a set of nodes and associated fields to create agents in Subsection \ref{sec:overview}. We then delve into the key concepts and mechanics of the AutoGRAMS frameworks, focusing on; nodes and node types in Subsection \ref{sec:node types}, transitions in Subsection \ref{sec:transitions}, variables and memory in Subsection \ref{sec:variables}, functions in Subsection \ref{sec:functions}, methods of designing autograms in Subsection \ref{sec:design}, and finally, autogram configurations and settings in Subsection \ref{sec:config}. Some other details of the framework are included in the appendix, including  how prompts are formed (Appendix \ref{appx:prompts}), and details of how language models are used (Appendix \ref{appx:language-models}), a description of interjection nodes that help model unexpected user replies (Appendix \ref{appx:interjection}), and simulating user replies in conversational autograms (Appendix \ref{appx:user}),

\subsection{General overview}
\label{sec:overview}

AutoGRAMS allows for a unified representation between graphical chatbots that use states and transitions, general code, and approaches that leverage LLMs internal reasoning abilities. An AutoGRAMS program, which we refer to as an autogram (autonomous program)~\footnote{An ``autogram'' can also refer to a sentence that describes itself by providing an inventory of its own characters. Likewise, autograms in this framework are also self-referential--they contain a reference to their own object that can be used to self-modify.}, is defined by a collection of nodes and the transition functions between them. Within the nodes, there are associated fields that define instructions to be executed and transition behaviors/functions to other nodes.  

\begin{figure}[h]
\begin{center}

\includegraphics[width=\textwidth]{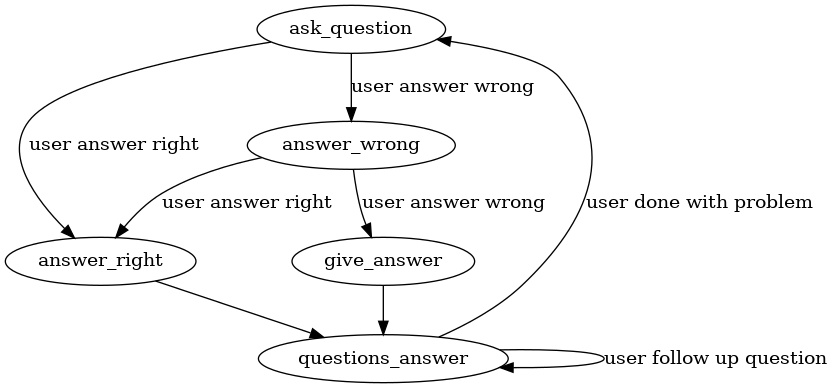}
\end{center}
\caption{Graphical representation of a chatbot created using an autogram with only chat-type nodes. Nodes represents a point where the language model will give a response, and each edge represents a different type of user response. This particular chatbot models an AI tutor that repeatedly quizzes the user about a subject. The chatbot starts with the \texttt{ask\_question} node where it asks the user a question, and then decides which node to visit next depending on whether the user's answer is right or wrong. If the answer is wrong, the interaction goes to the \texttt{answer\_wrong} node where the chatbot lets the user know the answer is wrong and asks them to try again. If the user answers wrong twice, the chatbot gives the answer. Eventually the chatbot reaches a node where it repeatedly answers questions the user has about the original question, and when the user is done asking questions, the chatbot revisits the original \texttt{ask\_question} node to ask another question. Each node executes a specific planned step in the conversation. Each node has a separate instruction for how the model should reply. Each node with multiple outgoing edges contains a multiple choice question that the language model uses to determine which node to visit next. So for instance, in this example, the \texttt{ask\_question} node has a transition question of \textbf{``is the user's answer correct? A. Yes B. No''}. If after viewing the user's reply, the language model predicts the answer is Yes, then the autogram will go to the \texttt{answer\_right} node to get the next instruction. Otherwise it will go to the \texttt{answer\_wrong node}.
}
\label{fig/trajectories}
\end{figure}

An autogram is represented by a set of nodes and their associated fields. The nodes are used to form a data structure representing a graph, and execution of the autogram consists of taking a trajectory of one or more steps along this graph. In a conversational setting, one of the main goals is to model different conversational trajectories, where each node represents a chatbot's reply, and each edge represents a user's reply. For instance, consider the conversational graph in Figure \ref{fig/trajectories}--An AI agent designed to quiz the user on a subject may start by asking a question, and depending on whether the answer is correct, will continue by either going over the answer with the user, or asking them to try again. If a user repeatedly answers incorrectly, the agent may provide an explanation and then proceed to a new question after addressing any further inquiries.  This type of agent can be modeled as a set of nodes, each with a unique set of instructions and transition behaviors. 

There are many possible fields than can be defined for each node, but some of the most important fields are:

\begin{itemize}
    \item \textbf{instruction:} This field determines the node's behavior and is interpreted differently depending on the node type. The instruction could be a language model prompt, executable code, or a call to another AutoGRAMS function.

    \item \textbf{action:}  This field determines the node type and how the \textbf{instruction} is processed. In simpler scenarios, such as Figure \ref{fig/trajectories}, all nodes can be designated as \texttt{chat} type, indicating that each node generates a response.

    \item \textbf{name:}  This provides a unique identifier for the node, allowing other nodes to reference it during transitions.

    \item \textbf{transition\_question:} This is a question posed to the language model to help determine the next node to transition to. Common examples include \textbf{``What did the user say?''} or \textbf{``Does the user want X?''}

    \item \textbf{transition\_choices:} This is a multiple-choice list of potential answers to the \textbf{transition\_question}. For example, it could be \textbf{[``The user said X``, ``The user said Y'', ``The user said Z'']}, \textbf{[``yes'', ``no'']}, or similar. The language model uses the \textbf{transition\_question} and conversation history to predict the most fitting \textbf{transition\_choices}.

    \item \textbf{transitions:} This is a list of node names that the agent can transition to, depending on the predicted answer to the \textbf{transition\_question}. For instance, if \textbf{transitions} is \textbf{[``node1'', ``node2'']}, \textbf{transition\_question} is \textbf{``Does the user want X?''}, and \textbf{transition\_choices} is \textbf{[``yes'', ``no'']}, then the autogram transitions to \textbf{``node1''} if the model predicts \textbf{``yes''} and to \textbf{``node2''} if it predicts \textbf{``no''}. The lists are order-dependent. 
\end{itemize}

Each node in chatbots like the one described in Figure \ref{fig/trajectories} performs the following steps:

\begin{enumerate}
    \item \textbf{Generate a Contextually Relevant Reply:}  This is achieved by passing node-specific instructions to an instruction-following language model, ensuring the reply aligns with the node's intended purpose.
    \item \textbf{Determine the Next Node:} In conversational autograms, this decision may depend on the user's new reply. To accomplish this, the designer may:
    \begin{itemize}
        \item Define a list of permissible next nodes within the graph structure.
        \item Define a multiple-choice question about the user's reply, with each answer choice corresponding to one of the possible next nodes. This question is then presented to a language model, and the predicted answer determines the node transition. 
    \end{itemize}
\end{enumerate}

Nodes come in different types as defined by their actions (see Section \ref{sec:node types}), and in conversational settings (e.g. chatbots), chat-type nodes pause to return a reply to the user, pausing the trajectory through the graph until the user replies. When the user replies, the autogram can decide which of the possible transitions to select based on the user's reply. So for instance, a node that provides a reply that asks a yes or no question may transition to a different node depending on whether the user answers affirmatively or negatively. This allows the designer of the autogram, which could be a person or itself an autogram, to pre-program proactive responses that lead conversations in well thought out ways towards objectives such as obtaining information from a user, troubleshooting a problem step by step, or completing a task.

Other types of nodes can perform other functions; e.g. use a language model to generate thoughts, call external APIs, or even execute arbitrary Python code, allowing for AutoGRAMS to be used for more general applications. Memory in an AutoGRAMS is set by variables (Section \ref{sec:variables}) defined at the execution of nodes, which can be referenced in later instructions or code executed by nodes at a future point. The AutoGRAMS framework is Turing complete with respect to its node definitions; the designer can implement nodes that result in loops and conditionals using graph logic, and  individual nodes can be configured to execute basic code statements and variable assignments. We show that most functional Python programs can be represented using an autogram, and we implement an AutoGRAMS compiler that can convert vanilla Python code (Section \ref{sec:compiler}). It also enables Python code and AutoGRAMS specific nodes to be easily integrated together. AutoGRAMS subgraphs can also be called as functions, allowing for AutoGRAMS modules to be reused (Section \ref{sec:functions}). An autogram can also reference its own object, allowing an autogram to be fully self referential and self-modifying (Section \ref{sec:self-mod}).

\subsection{Node types}
\label{sec:node types}

An autogram is defined by a set of nodes, where each node performs a single step of execution. An execution step could contain anything from calling a language model with a prompt to generate text, to executing arbitrary code. Each node has an instruction that determines what will be executed at that step. Each node also has an 'action' which defines what type of node it is, and how that instruction will be interpreted.

The main actions in AutoGRAMS, which also correspond to types of nodes, are as follows

\begin{itemize}
    \item \textbf{Chat actions:}
    \begin{itemize}
        \item \textbf{chat:} In these nodes, the instruction serves as a turn-specfic conversational prompt for the language model, incorporating the current memory state of the autogram. The autogram returns both the language model's response and the updated memory state, allowing the system to pause and wait for further user input.
        \item \textbf{chat\_exact:} This action bypasses the language model's response generation. Instead, it directly outputs the instruction as a pre-programmed conversational reply, providing more precise control over specific responses.
    \end{itemize}
    \item \textbf{Thought actions:} These nodes utilize the instruction as a text generation prompt for the language model. Unlike \textbf{chat} nodes, there is no pause for user input after execution; instead, the autogram immediately transitions to the next node. This type of node is useful for internal reasoning steps.
    \item \textbf{Python actions:} In this case, the instruction is treated as Python code and executed directly by the Python interpreter, enabling custom logic and functionality within the autogram.
    \item \textbf{Function actions:} These actions trigger the execution of a separate subgraph within the AutoGRAMS graph. The instruction contains a reference to a callable AutoGRAMS node (potentially with arguments) to initiate this subgraph, returning control to the calling node upon completion. (For more details, refer to Section \ref{sec:functions} on AutoGRAMS functions.)
    \item \textbf{Prompt actions:} This action type focuses on modifying the initial prompt presented to the language model. Language models receive both a turn-specific prompt (the instructions for chat and thought nodes) and an initial, overarching prompt. This action provides the capability to reset or alter this initial prompt, influencing the model's behavior across the entire conversation.
    \item \textbf{Transition actions:}  These actions are primarily structural and do not utilize the instruction field. They serve as placeholders within the AutoGRAMS graph, allowing for additional branching points and greater control over the flow of the conversation without requiring specific instructions.
\end{itemize}

\begin{figure}[h]
\begin{center}
\includegraphics[width=\textwidth]{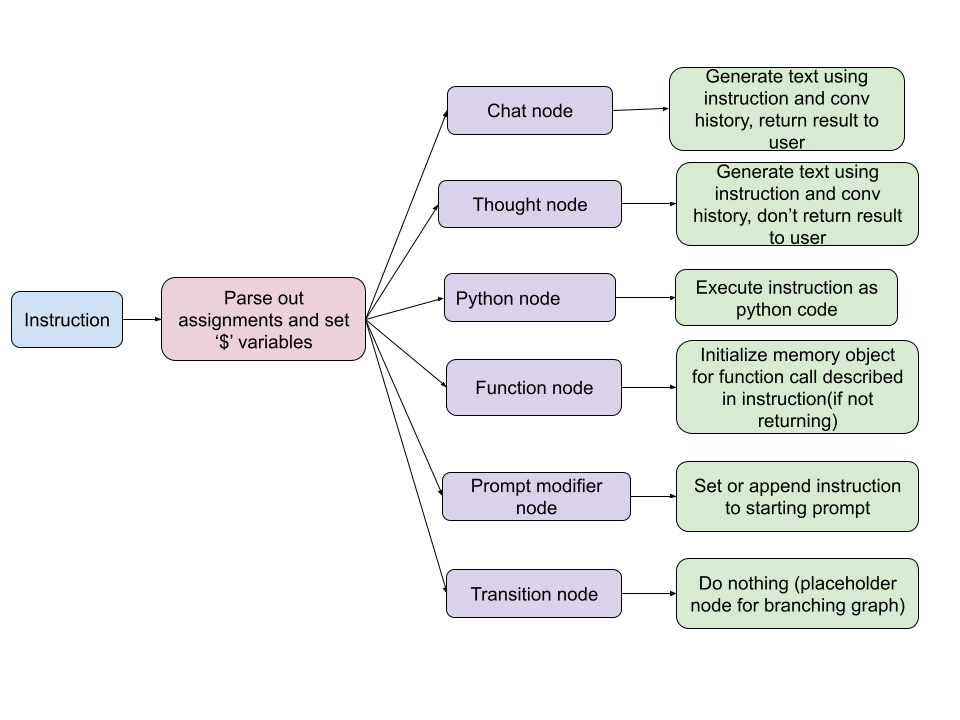}
\end{center}
\caption{Depiction of how a node's instruction is executed in AutoGRAMS. Behavior can vary greatly depending on the type of node. `\$' variables are covered in Section \ref{sec:variables}}
\label{fig:inst}
\end{figure}

We describe how AutoGRAMS instructions are executed depending on node type in Figure \ref{fig:inst}.

\subsection{Transitions}
\label{sec:transitions}

After executing the instructions within a node, the autogram must select the next node to execute. Each node contains fields that govern this selection process.

Each node has a \texttt{name} field and a \texttt{transitions} field, which stores a list of strings. In the simplest case, each string in \texttt{transitions} directly references the name of another node in the graph. A node with only one possible transition would list a single name, while nodes with branching possibilities would list multiple names.

In other cases, a string in \texttt{transitions} may not directly refer to a node name. For example:

\begin{itemize}
    \item \textbf{Wildcard Transitions:} A transition string with a ``.\*'' suffix (e.g., ``mynode.*'') indicates a conditional transition. This assumes the existence of nodes named ``mynode.a,'' ``mynode.b,'' etc., each with a defined \texttt{boolean\_condition} field containing a Python statement that may reference variables in memory (see Section~\ref{sec:variables} for more on Variables). If-elseif-else logic is then used to select the appropriate node based on the evaluation of these boolean conditions. (see Algorithm~\ref{alg:wildcard_transition} for a pseudo-code example, and Appendix~\ref{appx:wildcard} for a visualization)
    \item \textbf{Return Transitions:} A transition string of \textbf{``return''} (or \textbf{``return \textit{variable\_name''}} to return a variable) signifies a return from a function call within an AutoGRAMS function, transitioning back to the calling node.
    \item \textbf{Variable Transitions (advanced use case):} A transition string can also reference a variable, using its value in memory to determine the next node. This requires careful validation to ensure the variable's value corresponds to a valid node name. A variable transition could become a return or wildcard transition, since a variable could specify any string--which could correspond to any transition type.
\end{itemize}

Function-type nodes override this standard transition behavior. They transition directly to the node specified in their instruction, returning to the original node upon function completion. (See Section~\ref{sec:functions} for details on AutoGRAMS functions.)

\begin{algorithm}
\caption{\\ Pseudo-code showing how wildcard transitions implement if/else logic. Assume selected transition is ``mynode.*'', and there is a dictionary called nodes which contains nodes with names ``mynode.a'',``mynode.b'',``mynode.c''}
\label{alg:wildcard_transition}
\begin{algorithmic}[1]

\\
\If{nodes[``mynode.a''].boolean\_condition}
\\
\State $node \leftarrow$ nodes[``mynode.a'']

\ElsIf{nodes[``mynode.b''].boolean\_condition}
\\
\State $node \leftarrow$ nodes[``mynode.b'']
\Else
\\
\State $node \leftarrow$ nodes[``mynode.c'']
\EndIf
%
\end{algorithmic}
\end{algorithm}

\begin{figure}[h]
\begin{center}
\includegraphics[width=\textwidth]{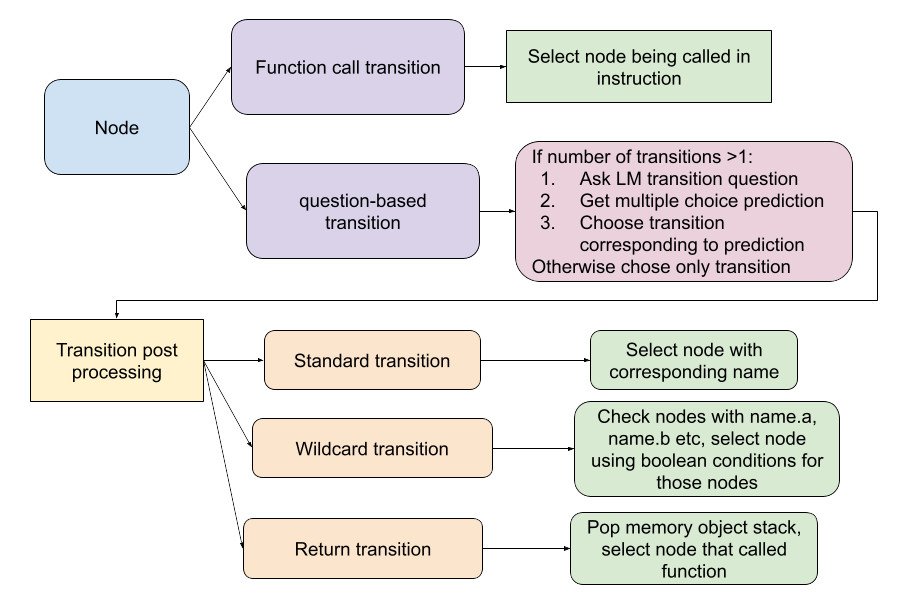}
\end{center}
\caption{Depiction of how the next node is selected from a previous one in AutoGRAMS.  If the previous instruction executed a function call, the typical transition behavior is overridden and the autogram transitions to the node called in the previous instruction. If the number of defined transitions for that node is more than one, then the nodes transition question and transition choices are used to construct a multiple choice question, which is passed to a language model. The language model's prediction determines which transition to select. If the selected transition is a variable, this is set during transition post-processing. If the selected transition corresponds to the name of an existing node (standard transition), than that node is selected. If the transition is a ``return'' transition, the previous function calling node is selected as the next node. Lastly, if the transition is a wildcard transition (ending in ``.*''), if/else-if/else logic is used to determine which of 2 or more nodes with the same prefix as the transition is selected. }
\label{fig:transitions}
\end{figure}

Figure \ref{fig:transitions} presents a flowchart of how nodes are selected from the previous node.

\subsection{Variables}
\label{sec:variables}

AutoGRAMS allows the use of variables to store and manipulate memory, enabling dynamic behavior within an autogram. Variables in AutoGRAMS are managed as Python variables and can be created, referenced, and modified throughout the execution of an autogram. 

\subsubsection{Variable Assignment and references}

Variables are assigned using the `=` sign within the instructions of nodes. The node's instruction (with the assignment parsed out) will be executed in accordance with the node's action, as previously described in Section \ref{sec:node types}. This means that the type of node will also influence what value is assigned to the variable. 
\begin{itemize}
    \item \textbf{Python Nodes}: Execute Python code directly. For example, a node with the instruction, 
    
    \texttt{x = [0, 1, 2]}
    
    initializes a variable \texttt{x} to a list.
    \item \textbf{Chat and Thought Nodes}: Use the language model to generate text. For example, an instruction like,
    
    \texttt{x = summarize the recent conversation history}
    
    will store the generated text in the variable \texttt{x}. In this specific example, the text stored in the variable \texttt{x} will be a language model summary of the recent conversation history. i.e. at run time, this instruction will be executed and its results stored as a Python string in the variable \texttt{x}.
    
    \item \textbf{Function Nodes}: When a function node is applied, it can pass variables as arguments to the called AutoGRAMS function. The calling node’s instruction specifies the variables being passed. Within the called function, these variables are assigned and can be used or modified. For instance, a function call, 
    
    \texttt{summary = summarize(text)} 
    
    will have the variable \texttt{text} accessible within the subgraph of the \texttt{summarize()} node. When that subgraph encounters a return transition, it will return a value that will be stored in the variable summary. See Section \ref{sec:functions} for more details on how this works.
\end{itemize}

Although it is technically possible to assign variables in other node types, it is not recommended and currently has no utility.

Variables can also be referenced in two ways:
\begin{enumerate}
    \item \textbf{Direct Use}: Variables are used directly in Python code within the instructions of Python or Function-type nodes, or in the \texttt{boolean\_condition} attribute of a node (used for wildcard-transitions). For instance:
    \begin{verbatim}
    sorted(list1)
    \end{verbatim}
    could be used as an instruction for a Python node to sort a list called \texttt{list1} that was assigned as a variable at a previous node.
    
    \item \textbf{\$ Variable Embedding (\$ syntax)}: Variables are embedded as strings within instructions or other node fields using \$ syntax. This is useful in chat and thought-type nodes. For example:
    \begin{verbatim}
    Here is a summary of what was previously discussed: $summary. 
    Reply to the user and remind them what was previously discussed.
    \end{verbatim}

\end{enumerate}

\textbf{\$ variable embedding} can also be applied in the \texttt{transition\_question} attribute of a node to allow the multiple transition question to change dynamically. They can also be used as transitions directly, although this requires care. For example, setting
\begin{verbatim}
transitions = ["node1", "node2", "$variable_node"]
\end{verbatim}
, for a particular node would allow the third transition to be dynamically dependant on the value of \texttt{variable\_node}, however this will not execute successfully if \texttt{variable\_node} is not a valid transition.

\subsubsection{Variable scopes}

The memory in AutoGRAMS is managed through a memory object, which stores variables and the conversation history. This object is structured as a stack, where each function call adds a new layer, and each return removes the top layer. If no functions are used, a variable can be accessed anywhere within an autogram. However, local functions are unable to view their calling scope, including any variables in that calling scope. See Section \ref{sec:functions} for more details.

\subsubsection{Example of Variable Usage}

Consider a set of nodes that demonstrate variable assignment and referencing:

\begin{verbatim}
name: "write_topics"
instruction: "topics = List the topics covered in today's lesson."
action: "thought"
transitions: ["append_topics"]
\end{verbatim}

\begin{verbatim}
name: "append_topics"
instruction: "all_topics.append(topics)"
action: "python_function"
transitions: ["tell_topics"]
\end{verbatim}

\begin{verbatim}
name: "tell_topics"
instruction: "Inform the user about the topics covered today, which were: $topics"
action: "chat"
\end{verbatim}

In this example:
\begin{itemize}
    \item The \texttt{write\_topics} node initializes calls a language model to write down a list of topics and store it in a variable called topics, which will be a string.
    \item The \texttt{append\_topics} node applies its instruction as Python code, appending the variable \text{topics} to a list called \texttt{all\_topics} that is presumed to have been defined previously.
    \item The \texttt{tell\_topics} node uses \$ variable embedding to dynamically insert the \texttt{topics} variable within the instruction string to get a response.

\end{itemize}

By utilizing variables, AutoGRAMS facilitates memory management and dynamic interactions, enabling the creation of sophisticated and adaptable AI agents.

\begin{figure}[h]
\begin{center}
\includegraphics[width=.6\textwidth]{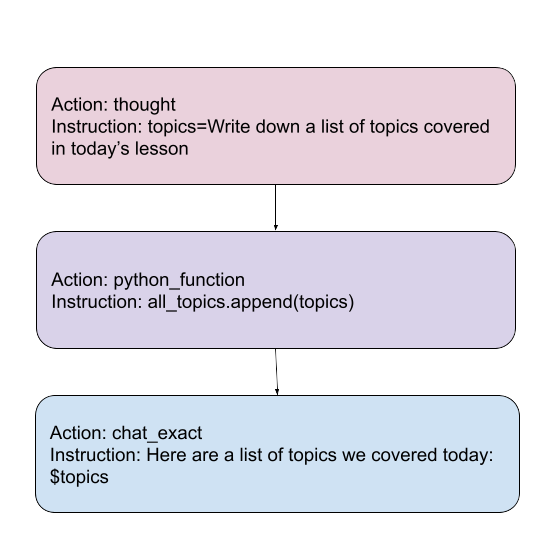}
\end{center}
\caption{Example of a set of nodes that use variables. The exact behavior of variable assignment can depend on the type of node, but for thought and chat style nodes, this is the text generated by the language model, and for python-type nodes this is the result of the code execution using the Python interpreter. Variable assignment allows for memories to be stored that can be referenced in future nodes. Variables can be referenced in 2 ways--directly or using \$-syntax. Direct use means using a variable as is, and can only be used in fields that will be passed to directly the Python interpreter. This includes the instruction of python-type nodes (see the $python\_function$ node above), and the $boolean\_condition$ attribute of a node that is often used in combination with wild card transitions. \$-syntax means embed the variable as a string on the fly; for instance in the example above, the model will reply with the exact text set in the topics variable inserted into the instruction (a $chat\_exact$ node replies with the exact instruction without calling the language model). \$-variable references are most straightforwardly used in $instruction$, $transition\_question$ to facilitate on-the-fly instructions to a language model or on-the-fly transition questions that incorporate a memory. With care, \$-variable references can also be used in the instruction of python-type nodes to create dynamically changing code, or as elements of the transition list to allow for dynamically changing transitions in the graph.}
\label{fig:variables}
\end{figure}

\subsection{Function calls and scopes}
\label{sec:functions}

\textbf{AutoGRAMS} introduces the ability to call graphical modules as functions, allowing subgraphs to be called similarly to functions in programming. This enables modularity, reusability, and control over variable scopes.

\subsubsection{Function Nodes and Calls}
Function nodes allow for the execution of separate subgraphs within the main AutoGRAMS graph. These nodes can call other nodes within the graph or external programs. The process involves:

\begin{itemize}
    \item \textbf{Defining the Function Node}: A node designated as a function node will have its instruction specify the function to be called, such as,
    
    \texttt{summary = summarize(document)}.
    \item \textbf{Passing Arguments}: Variables can be passed as arguments to the called function. Within the function, these variables are assigned and can be used or modified.
     \item \textbf{Defining a callable node}:
     A node can be made callable by its name. Node name with parentheses are callable. So a node named \texttt{summarize} cannot be called a s function, a node named \texttt{summarize()} can be called as as function with no arguments. A node named \texttt{summarize(text1,text2)} can be called with 2 arguments. The arguments specified by the node name determine what the variables will be named in the scope of the called node.
     \item \textbf{Defining the Subgraph of the callable node}:
     The callable node, which acts as the root node of the subgraph, can have transitions to other nodes that perform other operations. Transitions use the string ``return'' or ``return varname'' can be added to specify that the subgraph should end. 
     
    \item \textbf{Executing the Subgraph of the callable node}: The function node transitions to the callable node specified in the instruction. The autogram executes the subgraph of the callable node until it reaches a return transition, which may optionally return specific information back to the calling node.
\end{itemize}

\subsubsection{Return Transitions}
Return transitions mark the end of a function’s execution and return control to the calling node. They can also return specific values to be used by the calling node. For example, \texttt{return topics\_summary} will pass the value of \texttt{topics\_summary} back to the calling node.

\subsubsection{Scopes in Function Calls}
The scope of variables in function calls is crucial for maintaining the integrity of the memory and ensuring correct execution. \textbf{AutoGRAMS} supports different scope types for function calls:

\begin{itemize}
    \item \textbf{Local Scope}: The function can only access variables passed as arguments. Variables within the function do not affect the calling scope. This is specified using the \texttt{local\_function} action of the calling node.
    \item \textbf{Global Scope}: The function can access and modify all variables and conversation turns in the calling scope. This is specified using the \texttt{global\_function} action of the calling node.
    \item \textbf{Mixed Scope}: The function can read all variables and conversation turns from the calling scope, but the variable and conversation turns set during execution are erased after returning. This is specified using the \texttt{function} action of the calling node.
\end{itemize}

\subsubsection{Example of Function Usage}

Consider a set of nodes that demonstrate the use of function calls and scopes:

\begin{verbatim}
name: "call_summarize_docs"
instruction: "summary = summarize_and_combine(document1,document2)"
action: "local_function"
transitions=["process_summary"]
\end{verbatim}

\begin{verbatim}
name: "summarize_and_combine(text1,text2)"
instruction: "summary1=write a summary of the following text: $text1"
action: "thought"
transitions: ["summarize_second"]
\end{verbatim}
\begin{verbatim}
name: "summarize_second"
instruction: "summary2=write a summary of the following text: $text2"
action: "thought"
transitions: ["combine_summaries"]
\end{verbatim}

\begin{verbatim}
name: "combine_summaries"
instruction: "combined_summary=Write a summary combining 
            summary 1 -- $summary1 
            summary 2 -- $summary2" 
action: "thought"
transitions: ["return combined_summary"]
\end{verbatim}

In this example:
\begin{itemize}
    \item The \texttt{call\_summarize\_topics} node calls the \texttt{summarize\_and\_combine()} function using its instruction. The \texttt{document1} and \texttt{document2} variables are passed as arguments.
    \item \texttt{summarize\_and\_combine()} function starts at the node named \texttt{summarize\_and\_combine(text1,text2)}. The variable \texttt{document1} in the calling namespace will be mapped to text1 in the called namespace, and the variable \texttt{document2} in the calling name space will be mapped to \texttt{text2} in the called namespace. 
    \item After the function call, the autogram will execute the \texttt{summarize\_and\_combine()} node, followed by the \texttt{summarize\_second} and \texttt{combine\_summaries} nodes. 
    
    \item After executing the \texttt{combine\_summaries} node, it encounters a return transition specifying \texttt{return combined\_summary}. This will cause the autogram to return to the calling node, which was \texttt{call\_summarize\_docs}. The variable \texttt{summary} defined by the instruction of \texttt{call\_summarize\_docs} will be set equal to the returned variable \texttt{combined\_summary}.

    \item after the return is complete, the autogram will transition to to a node called \texttt{process\_summary} (not defined above) specified by the transitions of the \texttt{call\_summarize\_docs} node.
\end{itemize}

\subsubsection{Visualization of Function Calls}
\begin{figure}
\centering
\begin{subfigure}[b]{1\textwidth}
  \includegraphics[width=1\linewidth]{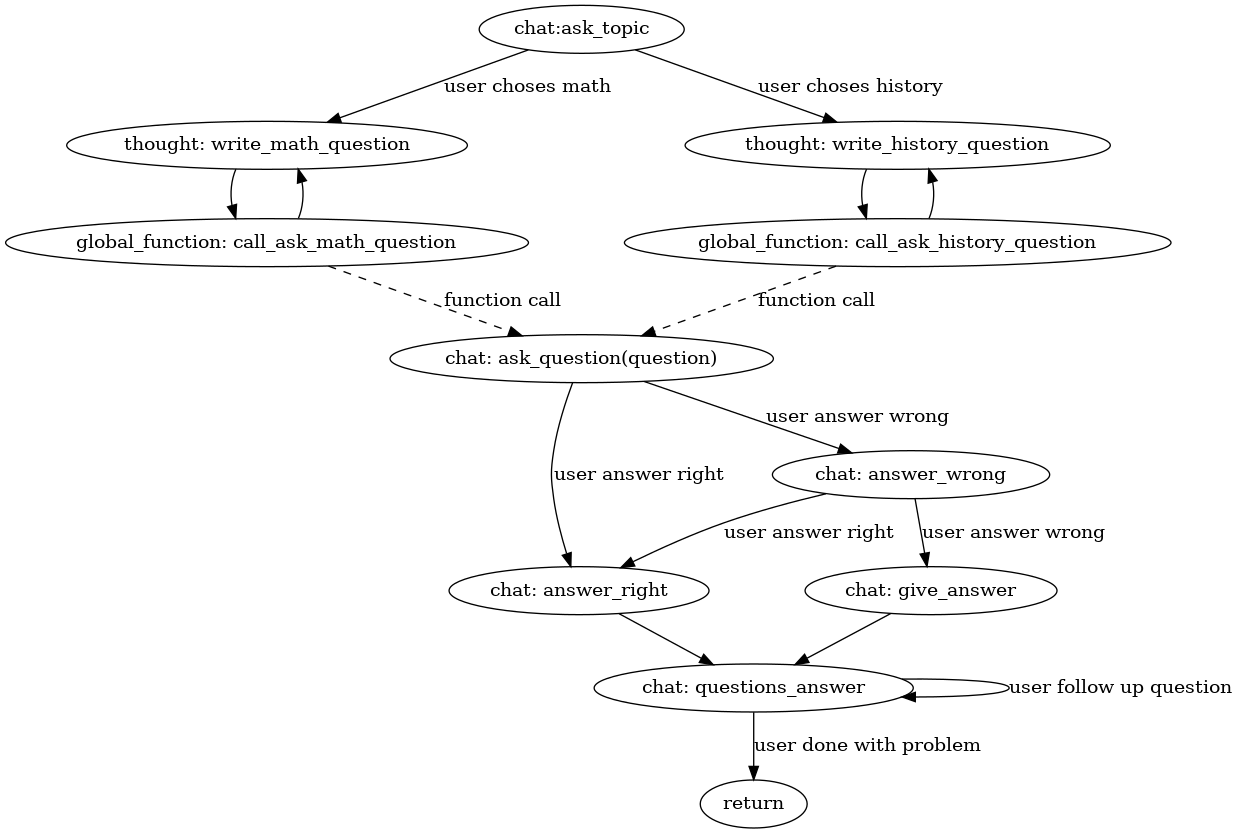}
   \caption{An autogram where multiple branches use a function call to the same conversational module to ask a question. Dashed lines represent function calls. When the return node is encountered, the autogram will transition to either \texttt{call\_ask\_math\_question} or \texttt{call\_ask\_history\_question}, depending on which node called it. After returning, the autogram will go to either \texttt{write\_math\_question} or \texttt{write\_history\_question} and repeat the process. For AutoGRAMS functions, the variables passed are specified in the calling node's instruction and are available within the function's scope using the name of the called node. The node's name must include parentheses and the appropriate arguments to be callable. The \texttt{global\_function} action defined in the calling node specifies that conversation turns that occur during the function call will still be visible after returning.}
   \label{fig:fun} 
\end{subfigure}

\begin{subfigure}[b]{1\textwidth}
   \includegraphics[width=1\linewidth]{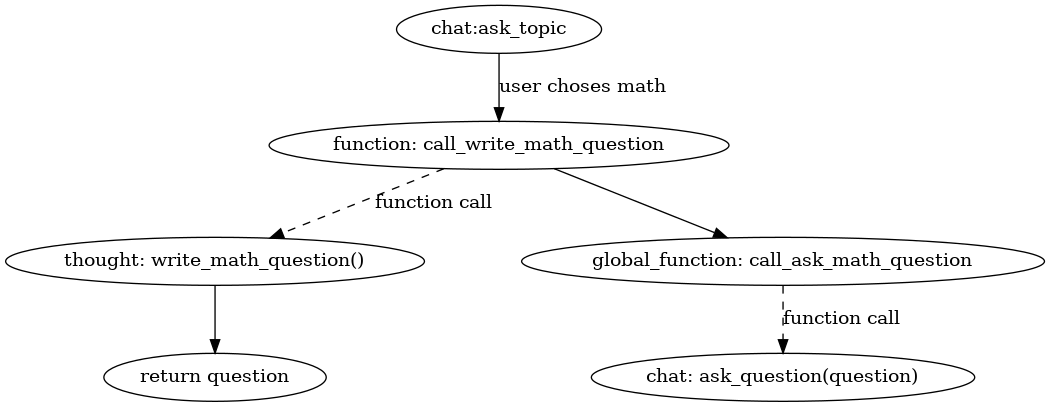}
   \caption{AutoGRAMS functions can also be utilized for non-conversational modules to enable module reuse or scope control. For example, setting the node action to \texttt{function} for \texttt{write\_math\_question()} allows the node to access past conversation history without permanently altering the stored conversation turns. The use of this scope with functions containing thought-type nodes prevents unnecessary additions to the permanent conversation history, allowing for more precise prompt control and reducing potential noise. The resulting question can be stored in a variable for later use. Additionally, since AutoGRAMS functions can be called externally from Python, they are suitable for non-conversational applications where the only external inputs are Python variables that the function can accept.}
   \label{fig:fun2}
\end{subfigure}

\caption{Depiction of function calls in AutoGRAMS. (a) depicts a function with chat-style nodes that allows the conversation to return to the calling node when a function returns. (b) depicts a function with thought-style nodes, which is useful for reusing modules and controlling scopes. AutoGRAMS functions can be called from within an autogram or from Python using the \texttt{autogram.apply\_fn()} method.}
\end{figure}

Figures \ref{fig:fun} and \ref{fig:fun2} illustrate how function calls are managed within \textbf{AutoGRAMS}, including the transitions and scope handling.

\subsection{Methods of Implementation and Visualization}
\label{sec:design}

At the time of writing, there are three ways to implement an autogram:

\begin{enumerate}
    \item \textbf{Spreadsheet-Based Design}:
    \begin{itemize}
        \item Each row in the spreadsheet corresponds to a different node in the AutoGRAM.
        \item Each column corresponds to a different node field, determined from the heading of the column.
        \item This method allows autograms to be defined and managed using a familiar spreadsheet interface.
    \end{itemize}
    
    \item \textbf{Pure Python Implementation}:
    \begin{itemize}
        \item Nodes of an autogram can be defined in Python by initializing an autogram object and using the \texttt{autogram.add\_node()} method to create new nodes with fields corresponding to the arguments provided.
        \item This is the most direct way to design an autogram, as all other methods of design are mapped to this.
        \item The main limitation is that it can be inconvenient for autograms that need to execute Python code within the autogram. Python code can be passed in as a string instruction, but this makes the code more difficult to read.
        \item Standard conditionals and loops can theoretically be implemented using wildcard transitions and loops in the graph with exiting branches, but these can be inconvenient to define graphically.
    \end{itemize}
    
    \item \textbf{AutoGRAMS Compiled from Python}:
    \begin{itemize}
        \item This method works best for autograms deeply integrated with Python code.
        \item It is technically a new language that closely relates to Python but has some differences that allow Python code to be integrated with AutoGRAMS nodes using Python syntax.
        \item AutoGRAMS is general enough to execute most functional Python programs using a combination of Python and transition-type nodes along with the right graph structure and wildcard transitions for branching.
        \item AutoGRAMS compiled from Python applies this conversion automatically, while also allowing autogram nodes with non-standard Python behaviors to be defined using a special function called \texttt{exec\_node}, which can also be written in the code to define where in the code those nodes should be applied.
    \end{itemize}
\end{enumerate}

\begin{figure}[h]
\begin{center}
\includegraphics[width=1\textwidth]{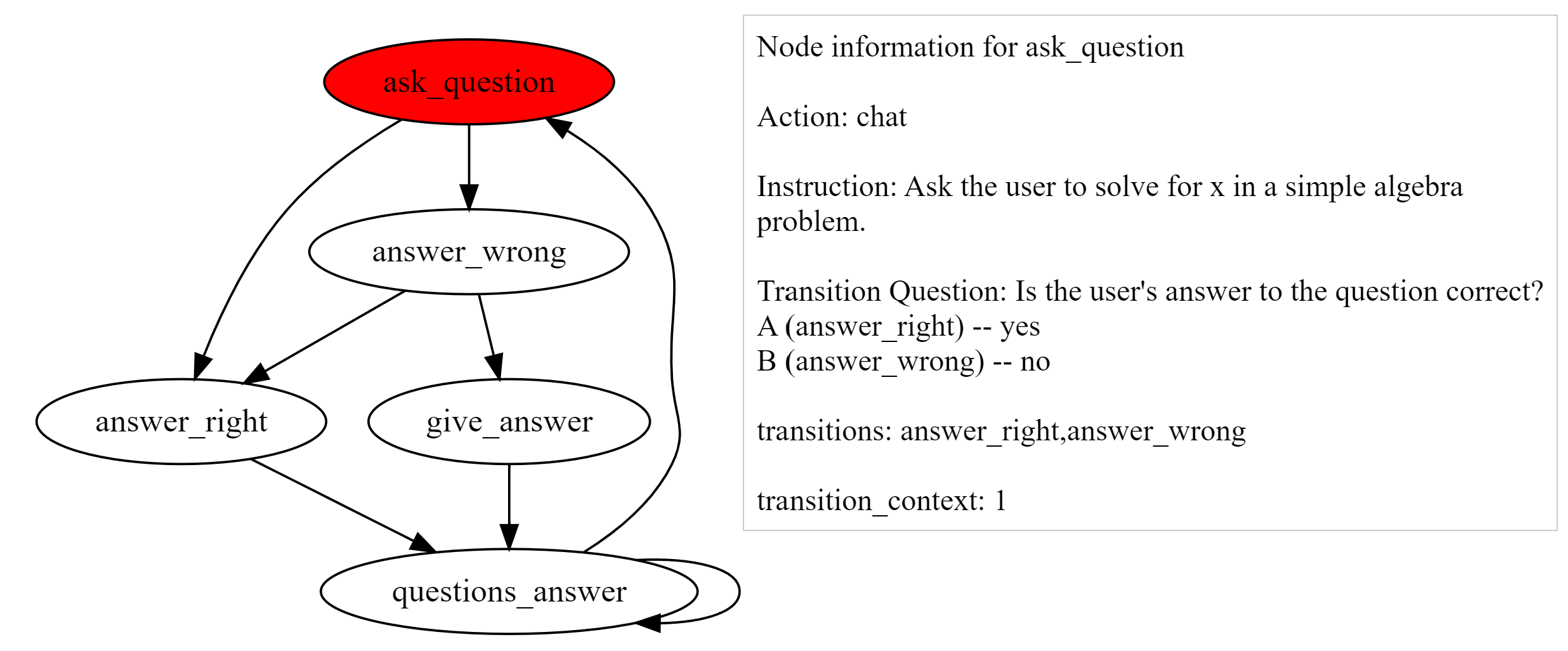}
\end{center}
\caption{AutoGRAMS can also be used to generate interactive graphs that allow all the fields associated with each node to be viewed by clicking on the appropriate node. It can be useful to periodically re-compile and view the graph while implementing an autogram. It is also possible to define node categories to allow a subset of the graph to be visualized. It is theoretically possible for any autogram to be implemented completely in a graphical interface such as the one above, and in the future we plan to implement graph-based AutoGRAMS editors that allow node fields to be changed and nodes to be added or removed.}
\label{fig:interactive}
\end{figure}

Any autogram can be fully visualized using a graphical interface. An interactive HTML file can be automatically generated for any autogram, displaying all the fields associated with each node. By clicking on a node, the code for that node is shown in a panel to the right. For more complex autograms, it is possible to specify subcategories of nodes to visualize, making the graph easier to understand, provided the category attribute is used in the nodes defined in the autogram. An interactive graph of an autogram is shown in Figure \ref{fig:interactive}. Future versions of AutoGRAMS will enable users to directly design and edit autograms from a similar interface.

\subsection{AutoGRAMS Configurations}
\label{sec:config}

AutoGRAMS configurations are essential for controlling various settings of an autogram. Key aspects of these configurations include:

\begin{itemize}
    \item \textbf{Model Selection}: Defining which models will be used within the autogram.
    \item \textbf{Prompt Templates}: Establishing templates to guide the language model responses.
    \item \textbf{Python Imports and APIs}: Specifying the Python modules and APIs accessible from the AutoGRAMS code.
\end{itemize}

Configurations are typically managed through a JSON file, with the exception of the Python modules field. This field requires actual Python modules containing code, which are passed to the autogram, allowing the use of any Python-defined function within Python function nodes.

For further details:
\begin{itemize}
    \item \textbf{Appendix \ref{appx:prompts}}: Provides an overview of how prompts for language models are set.
    \item \textbf{Appendix \ref{appx:language-models}}: Details the language models used in AutoGRAMS and their configuration settings.
\end{itemize}

These settings ensure that the autogram operates correctly, utilizing the appropriate models, prompts, and Python functionalities.

\section{AutoGRAMS graph compiler}
\label{sec:compiler}

AutoGRAMS gives the ability to embed general code within graphical AI agents and chatbots. One way to design autograms  is to define nodes one-by-one, with transitions to other nodes specified in node definitions. This is especially useful for chatbots, since conversations often have certain states that can be modeled as a graph or tree. AutoGRAMS also allows nodes in this graph to execute code, rather than give a reply. However, this requires the code that executes at a node to be passed as a string in a node's instruction. This graphical representation also means that loops and conditionals in the AutoGRAMS graph need to be defined using the appropriate transitions. We were motivated to allow deeper integration between general code statements and the graphical representations used by autograms. To do this, we implemented the AutoGRAMS compiler, which can map a combination of Python code and statements that define AutoGRAMS nodes, to an autogram that can be executed by the AutoGRAMS interpreter (Section \ref{sec:interpreter}). It does this by mapping the code in a file to a set of nodes with the appropriate attributes to perform the computation specified in the code file.

\begin{figure}
\centering
\begin{subfigure}[b]{0.6\textwidth}

   \includegraphics[width=1\linewidth]{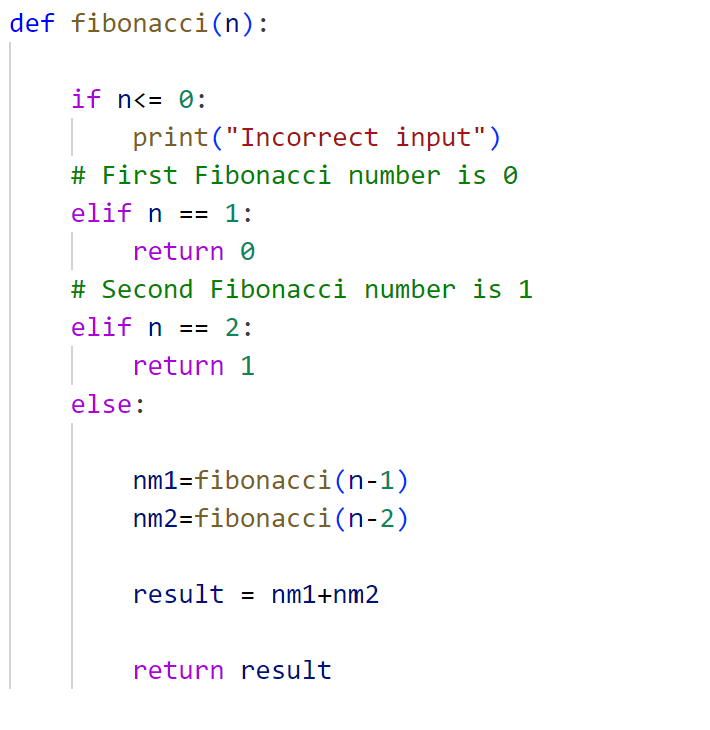}
   \caption{Python code for computing nth Fibonacci sequence number recursively. Functional Python code can be compiled into an AutoGRAMS graph automatically, with a few limitations on statements that call multiple functions or try to assign multiple variables.}
   \label{fig:Ng1} 
   
\end{subfigure}

\begin{subfigure}[b]{0.75\textwidth}

   \includegraphics[width=1\linewidth]{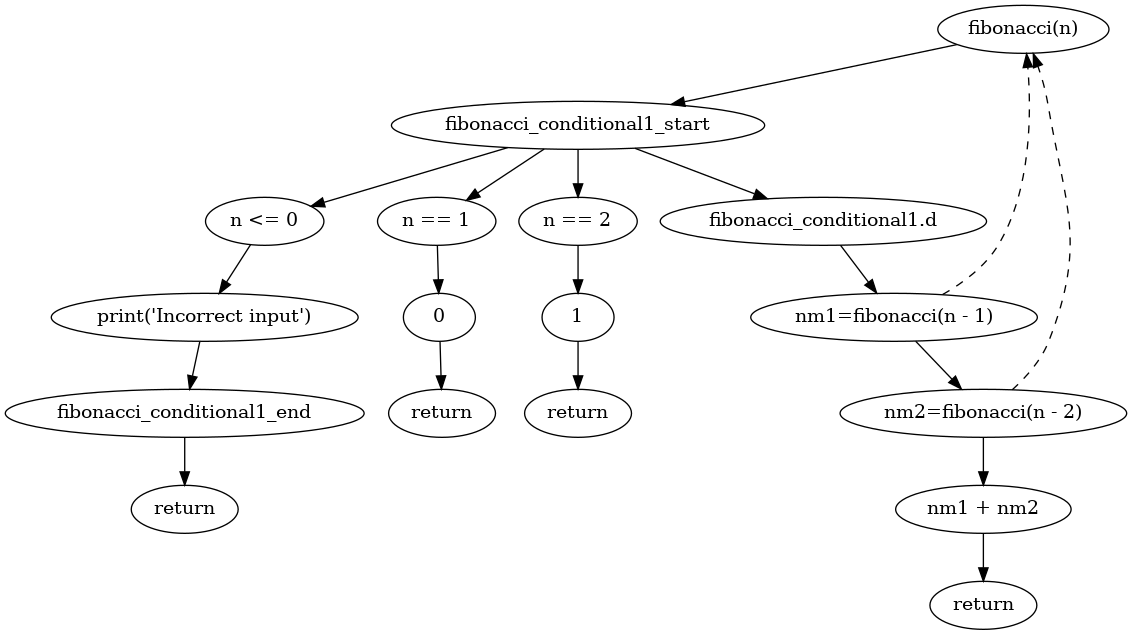}
   \caption{Automatically compiled AutoGRAMS graph for the code in (a). Code elements such as loops, conditionals, and function calls (including recursive function calls) can be handled by the autograms compiler, traverses the Python abstract syntax tree to compile an autogram from Python code. When the above autogram executes, the graph will follow the same trajectory as the program in (a) and yield an equivalent result. The AutoGRAMS compiler also allows other node types that use standard AutoGRAMS transitions to be interleaved with normal Python code, allowing Python coding elements to easily be conveniently be combined with any autogram.}
   \label{fig:Ng2}
\end{subfigure}

\caption{Mapping from (a) Python code to (b) AutoGRAMS graph computed automatically by the AutoGRAMS compiler. With a few exceptions that are not yet implemented, functional Python code can be mapped to an autogram directly by compiling the AutoGRAMS graph using the loops, conditions, functions, and variable assignments in the code. The AutoGRAMS compiler also allows for Python code, which will be mapped to python-type nodes, to be interleaved with AutoGRAMS thought and chat-type nodes. It also allows for AutoGRAMS functions and variables assignments to be written with similar syntax to Python}
\end{figure}

AutoGRAMS is general enough to execute functional programs that can be represented common features such as loops, conditionals, functions and variables. This is possible by combining python-type nodes with the desired flowchart needed to execute a program. For instance, consider a program that computes the nth element Fibonacci sequence recursively (While this is inefficient, we use it for illustrative purposes). Python code for this is given in  Figure \ref{fig:Ng1}.

The function $fibonacci(n)$ accepts a number, returns 0 of the number is 1, returns 1 if the number is 2, and returns the sum of $fibonacci(n-1)$ and $fibonacci(n-2)$ if the number is greater than 2. this function can be represented completely in AutoGRAMS with a combination of python-type nodes, function-calling nodes, transition nodes, return transitions, and wild card transitions. The first node of the graph must be a callable node named ``fibonacci(n)''. The parentheses in the name means that it can be called function calling nodes, and using n means that the first argument passed can be referenced with the variable name ``n''. A node with an action $local\_function$ and instruction of $x =  fibonacci(n)$ can be used to call this function as set the result to a variable called ``x''. 

The first conditional of the program is implemented with a wildcard transition, where the root node of the conditional applies a transition to $fibonacci\_conditional.*$, which means that the next node will either be $fibonacci\_conditional.a$, $fibonacci\_conditional.b$, etc. In this specific case, since there are 4 possible branches, it is necessary to define nodes named $fibonacci\_conditional.c$ and  $fibonacci\_conditional.d$. Each of these nodes has a $boolean\_condition$ attribute that can be used to specify the condition at which this branch can be executed. Boolean conditions are passed to the Python interpreter in AutoGRAMS so they can include any Python code or references to any variables visible at the current scope. $fibonacci\_conditional.d$ does not need a boolean conditional because it is the last node in alphabetical order. If the $n==1$ or $n==2$ condition are reached, the appropriate value can be returned by using a python-type that simply includes the desired number, and then applying a return transition immediately after, which returns the result of the previous node if no return variable is specified. If the else condition is met reaching the node $fibonacci\_conditional.d$, the autogram can use two $local\_function$ nodes that call the Fibonacci graph again with $n-1$ and $n-2$, sum the result, and return the result. The AutoGRAMS graph, which was compiled automatically from the Python code using the AutoGRAMS compiler, is given in Figure \ref{fig:Ng2}.

The AutoGRAMS compiler is used to convert code with python-like syntax into a set of AutoGRAMS nodes that can be executed as an autogram. It is also possible to code any autogram in pure Python (or a spreadsheet) by defining nodes one by one, in which case the AutoGRAMS compiler is not used. However, for autograms that contain Python statements or programmatic features like loops and conditionals, the compiled version has several advantages:

\begin{itemize}
\item Python statements can be included directly as code instead of as strings in the instruction of python\_function nodes

 \item Python style variable assignments can be be used instead of assignment variable in an instruction

\item AutoGRAMS function calls and functions can be defined using Python-like syntax. Note that functions are by default treated as local AutoGRAMS functions, but this behavior can be changed using the function decorator `@global\_function'' or ``@function'' above the function definition. 

\item the order of node chains to be inferred automatically from the order they appear in code, even if names and transitions aren't defined.

\item Python loops and if/else conditionals to compile into a graph automatically
\end{itemize}

Compared with standard Python programs, AutoGRAMS compiled from Python allows for \texttt{exec\_node()} statements to define AutoGRAMS nodes at any point in the code. These nodes can connect with other nodes defined by \texttt{exec\_node()} by specifying those nodes in the transitions argument, in which case these transitions will be governed by multiple choice questions asked to the language model. Chat type nodes defined in \texttt{exec\_node()} temporarily return a serializable object defining the entire state of the autogram (see Section \ref{sec:memory}), allowing a conversation to continue from a specific place in the code after receiving a new user reply. The graphical representations used by AutoGRAMS also give it certain advantages for self-modifying code, as discussed in Section \ref{sec:meta-self}.

The AutoGRAMS compiler recursively traverses the Python abstract syntax tree representation of the code to add AutoGRAMS nodes to the graph. A overview of how the Autograms compiler converts code into AutoGRAMS nodes is given in Appendix \ref{appx:compiler}.

\section{AutoGRAMS interpreter}
\label{sec:interpreter}
\begin{figure}[h]
\begin{center}
\includegraphics[width=\textwidth]{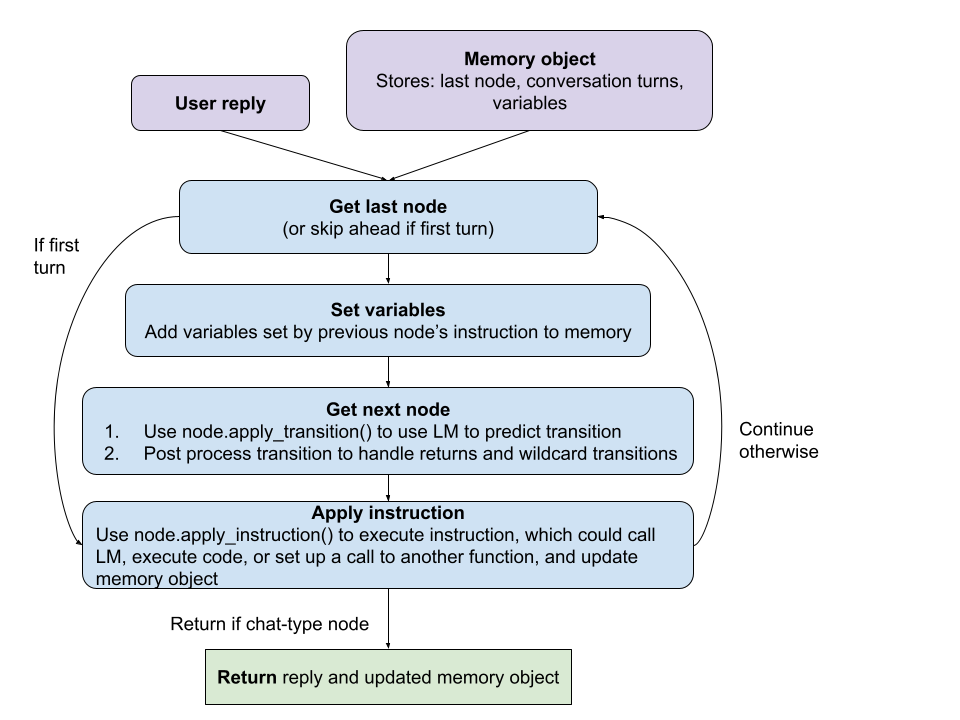}
\end{center}
\caption{Illustration of the core outer loop of the AutoGRAMS reply method. An autogram takes in a user reply and a memory object that models the full state of the autogram. For at least one iteration, the autogram selects the most recently executed node as saved in the memory object, sets any variables that were assigned at that node, and applies that node's transition function to select the next node. These first steps are skipped if no nodes have been executed yet and/or if the memory object is empty, in which case a designated start node is selected. The instruction of whichever node is executed, where the instruction may call a language model, be used to execute code or call an external Python API, or perform a number of other actions depending on the action defined for the node. If the node is a chat-type node, it will return a reply to the user. The autogram will continue to propagate along a trajectory through the graph until it encounters a chat-type node. }
\label{fig:autograms_loop}
\end{figure}

Autograms are executed via an iterative process that selects nodes and executes them. Within the AutoGRAMS framework, an autogram can be applied using the \texttt{reply} method, which is for conversational modules, and the \texttt{apply\_fn} method, which allows an AutoGRAMS function to be called directly from Python, and it meant for non-conversational modules. Both work in a very similar fashion, other than the way they terminate their main loop. The reply method terminates and returns a result when it encounters a chat node, and is designed to restart from the node it left off at when it receives another user reply. The \texttt{apply\_fn} method terminates when it encounters a return statement--meaning the AutoGRAMS function is done executing and is returning a result. An illustration of the outer loop of the AutoGRAMS reply method is given in Figure \ref{fig:autograms_loop}.

 \subsection{Main steps for obtaining conversational replies}

The outer loop of autogram.reply() has 6 main steps:

\begin{enumerate}

\item  get the variable output of previous node (skip this step if no previous node). This step calls a node specific method called \texttt{get\_variable\_output()} which returns the variable output of the node. The specific behavior of this will depend on the node type, but this is often the text generated by the chatbot.

\item  assign variables assigned in previous node to memory (skip this step if no previous node). If any variable outputs were in the previous node's instruction, the memory object (Section \ref{sec:memory}) assigns the nodes variable output to a variable with the name defined in that instruction. This goes in the top level of the memory objects stack.

\item  apply transition function from previous node  (skip this step if no previous node). This calls a node specific method called \texttt{apply\_transition()} to get an unprocessed new node id. The behavior of this method will can be different for different types of nodes, but is generally similar for non-function calling nodes. If there is only 1 possible transition in the \texttt{transitions} list of the node, then the result will usually this transition.If there are multiple transitions, the transition will depend on what the classifier predicts.There also may be interjection transitions if the node is a chat type. If the node is calling a function, the new node id will be the root node being called.

\item  post-process the new node id (skip this step if no previous node). If the new node id output by \texttt{apply\_transition()} corresponds to another node in the graph, and the next step will be to simply select that node. However, if the new node id corresponds to a return statement  or a wildcard transition, additional post processing will be needed. A return statement will require using the memory object to find the previous function calling node, and transition back to this node. A wild card transition requires finding the nodes with the matching prefix and applying if/else logic to determine which node to select.

\item  get the new node and apply its instruction. This calls a node specific method called $apply\_instruction()$ to execute the instruction of the node. This can very greatly for different node types. Apply instruction will always strip any variable assignments on the left of the node's instruction string. It will then render the string for any \$ variables in the instruction. The $apply\_instruction()$ for chat and thought type nodes will then call the language model with the instruction and conversation history used in the prompt. In python-type nodes, the instruction will be executed as code by the Python \texttt{eval} function, with all in scope variables and Python modules set by the autogram config included as local variables. 

\item  if the new node is a chat node, return a result with the models reply and the memory object storing the state of the program, continue the loop otherwise
  
\end{enumerate}

\subsection{Managing Memory}
\label{sec:memory}

\begin{figure}[h]
\begin{center}
\includegraphics[width=\textwidth]{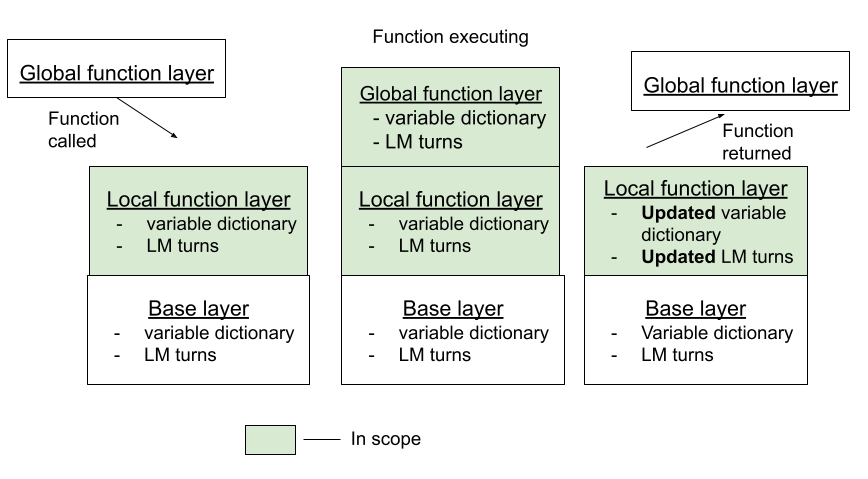}
\end{center}
\caption{Illustration of the memory object stack during 3 different points-1. the calling, 2. the execution, and 3. the returning of a function of type \texttt{global\_function}. Each layer of the memory object's stack stores a dictionary of variables assigned and the language model (LM) turns executed at that layer. When a function is called, a new layer is added to the stack. When variables or LM turns are retrieved during the execution of nodes, the memory object retrieves all previous turns and variables until it encounters a stack layer corresponding to a function of type \texttt{local\_function}, which cuts off access to lower layers of the stack. When the layer corresponding to a \texttt{global\_function} returns, all variables and LM turns are added to the calling layer, which in the above example is a \texttt{local\_function} layer. }
\label{fig:autograms_mem}
\end{figure}

 AutoGRAMS uses a ``memory object'' to keep track of the state of the program, including all the variables set by AutoGRAMS nodes and conversation turns. The memory object stores a stack of all of the layers of functions in an autogram. Each function call adds a layer to the stack, and each function return removes a layer from the stack. The type of function determines which variables and language model conversation turns are considered to be in scope at a certain layer of the stack. Within a global function or regular function, variables and conversation turns at the calling layer are visible, whereas in a local function these are not visible. Methods for obtaining variables and language model conversation turns within the memory object class add all variables or turns until they encounter a layer corresponding to a local function, which cuts off access to everything below the local function in the stack. When returning from a function, a layer in the stack is deleted. If the function was a global function, all variables and conversation turns are added to the calling layer. Otherwise these fields are erased, and only variables modified or returned by the function will be remembered. The stack of the memory object is illustrated in Figure \ref{fig:autograms_mem}.

 In addition to storing the stack, the memory object is also used to log model turns and nodes visited. The entire state of an autogram can be recovered from the memory object, allowing autograms to continue from where they left off after executing chat-type nodes that wait for a user reply. The memory object and user reply can be passed to the autogram to obtain the model's reply.

 \subsection{Interpreting Python statements}

Python statements (as well as the `boolean\_condition` field for wild card transitions, and arguments to AutoGRAMS functions) are interpreted by the Statement Interpreter. At initialization, the Statement Interpreter uses the Autogram Config to load all Python imports and modules that will be allowed within the scope of the program. It also overrides any Python builtins that are not explicitly listed in the Autogram Config to prevent them from being called. These imports are effectively treated as global variables that can be accessed anywhere from within the autogram. When the Statement Interpreter called to execute code, it uses the Python \texttt{eval} command--it includes all variables in scope in the memory object, as well as Python modules and imports passed from the Autogram Config in the scope of the code execution during the call to \texttt{eval}.

\section{Self-modifying and meta autograms}
\label{sec:meta-self}
\subsection{Introduction}

AutoGRAMS allows a designer to specify a series of instructions and transitions that form a program that control this process. However, the space of possible useful processes like this that can be formed is extremely large. For instance, there are likely many processes in AutoGRAMS that would be useful for solving certain problems, or handling certain conversational scenarios. It could be difficult for the designer to specify every possible process that the agent may need to handle. It would therefore be very useful to have a ``meta-process'' that can design large sets of these processes in advance, or be able to define new processes on the fly when encountering new scenarios.

AutoGRAMS is designed in such a way that the meta-process can be of the same form (an AutoGRAMS graph) as the processes that the meta-process designs. This makes it possible to design meta-processes that can self-modify, giving greater flexibility as compared with a system where the meta-process is of a different form from the process. Within AutoGRAMS, we refer to these meta-processes as meta-autograms.

\begin{figure}[h]
\begin{center}
\includegraphics[width=1\textwidth]{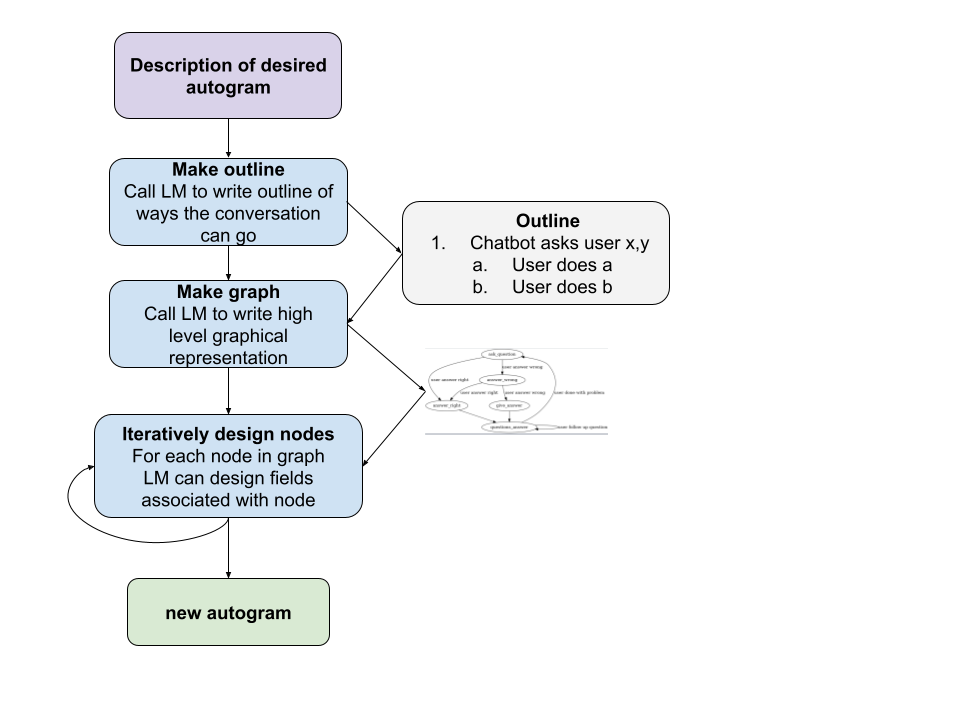}
\end{center}
\caption{An autogram can be used to design another autogram, which we refer to as a meta-autogram. For instance from a prompt. This diagram gives an overview of a meta-autogram we developed. A series of thought-type nodes can be used to first draw a high level representation of the new autogram's graph. After drafting this, the meta-autogram can extract the nodes in the new graph as a list, and iterate through thought nodes that write the fields for each node in the new autogram. The meta-autogram finishes by returning the arguments needed to create a new autogram based on the instructions fed in at the start of the meta-autogram. }
\label{fig:meta-autogram}
\end{figure}

\subsection{Meta autograms}

One of the goals of AutoGRAMS was to design an agent framework general enough that an agent could be used to design another agent of the same form. AutoGRAMS provides a simple working example of an AutoGRAMS function that can be used to design a chatbot autogram. The function, which we refer to as a meta-autogram, takes in a prompt as an argument. It applies the following steps

\begin{itemize}
\item apply a thought node that asks the language model to design an outline of the possible ways the conversation can go from this prompt

\item apply a thought node that asks the language model to design a graph using the dot (graphviz)~\footnote{https://graphviz.org/} language, where each node should be labeled by how the agent should reply, and each edge should be labeled by how the user should reply.

\item call an external Python module to parse the nodes from the graphviz graph, go back to step 2 if problems with the graph are detected

\item For each node parsed from the graphviz graph, call thought type nodes to generate the attributes associated with that node. The prompt for the thought node conditions on the graph labels for that node from step 2 to guide it. All the attributes are saved as variables and used to form a diction ary of arguments for each nodes

\item return the arguments needed to initialize a new autogram

\end{itemize}

This process is visualized in Figure \ref{fig:meta-autogram}.

\subsection{Self-modifying autograms}
\label{sec:self-mod}

One advantage of the graphical representation of AutoGRAMS is that a node at one point in the graph can perform operations that modify nodes at other points in the graph. For instance, a series of nodes could be configured to perform the following operations

\begin{enumerate}
    \item select a node in the graph
    \item call a language model to decide what the new attributes of that node should be, and save the arguments to define this node in variables
    \item if any new nodes need to be defined (because transitions of original node changed), loop through and define new nodes using the process in step 2. Continue to define child nodes until all nodes are connected to the graph (the system will eventually need to connect all nodes back to existing nodes so that it doesn't get stuck in an infinite loop).

    \item loop through to initialize/modify nodes using the arguments saved in variables during steps 2 and 3

\end{enumerate}

This is one of many possible ways an autogram could self-modify by using nodes to modify other nodes or adding new nodes.

We provided the ability for autograms to directly access their own data structure during their execution when running in ``self-referential'' mode. An autogram is implemented as an instance of the Autogram class in Python, and when self-referential mode is enabled in the autogram config, nodes are able to reference a variable called \texttt{self} which refers to the autogram's own object that interprets the autogram. This potentially allows any of the autograms nodes, methods, or other attributes to be accessed during the execution of the autogram. One potential usecase for this is dynamic modification of the autogram. For instance, it is possible for an autogram to add new node to itself by calling the \texttt{self.add\_node()} method with the appropriate arguments from a python-type node. We implemented a simple proof of concept of this where an autogram simply adds and executes a new node at each turn, instead of having the nodes defined before hand. This works by having an AutoGRAMS function to generate all the attributes of a new node and add it to the autogram, and then when the function returns, apply a dynamically defined transition with a variable name that specifies the name of the new node. The new node then transitions back to the node that calls the function to design a node, allowing the chatbot to continue indefinitely. The full set of nodes needed to perform this algorithm along with explanations are provided in Appendix \ref{appx:self-ref}.

\subsection{Future Directions}

Our future work will focus on developing software libraries of self-modifying autograms capable of defining useful autogram processes and learning from their interactions with the environment. In this paper, we demonstrated an autogram that can design an AutoGRAMS chatbot based on a prompt describing the chatbot's required functionality. While the scope of the present paper was to develop useful representations for designing such algorithms, future work will aim to implement autograms that perform these functions.

Future applications could include autograms that process conversations conducted by a human agent and generate new autograms to replicate the human agent's logic. Additionally, an autogram could be programmed to analyze its own interactions, identify where it went wrong due to unexpected user input, and modify its nodes accordingly.

Advanced autograms could also retrieve information on complex topics and adjust their reasoning structures to better integrate this material into their conversational or reasoning routines. This has significant implications for using large language models or generative AI in robotics. By fine-tuning language models to better model autograms, we hope to enhance the ability of autograms to design other autograms and self-modify.

\section{Related Work}
The development of AutoGRAMS is situated within a broader landscape of research encompassing natural language processing (NLP), AI agents, and programming languages for AI. This section reviews key advancements and their contributions to the field, highlighting how AutoGRAMS builds on and extends these innovations.

Significant progress in NLP and neural language modeling \citep{bengio2000neural} has been driven by transformer-based architectures \citep{vaswani2017attention}, such as BERT (Bidirectional Encoder Representations from Transformers) \cite{devlin2019bert}, GPT-3 (Generative Pre-trained Transformer 3) \cite{brown2020language}, and T5 (Text-To-Text Transfer Transformer) \cite{raffel2020exploring}. These models have shown exceptional capabilities in understanding and generating human-like text, providing a robust foundation for developing advanced conversational agents. Notable conversational agents include Google’s Meena \cite{adiwardana2020towards}, Facebook’s BlenderBot \cite{roller2020recipes}, and Microsoft’s DialoGPT \cite{zhang2019dialogpt}, which leverage large-scale pre-training and fine-tuning on diverse datasets to perform a variety of conversational tasks. AutoGRAMS relies heavily on the ability of pretrained Transformer models to generate realistic text that can be used in conversational replies and apply reasoning steps.

The intersection of programming languages and AI has been extensively explored to facilitate the integration and development of AI models. Python, along with frameworks like TensorFlow \citep{abadi2016tensorflow} and PyTorch \citep{paszke2019pytorch}, has become integral to AI research, providing comprehensive libraries for model development, training, and deployment. Domain-specific languages (DSLs) like Keras for neural network modeling \citep{gulli2017deep} and Rasa for building conversational AI \citep{bocklischrasa} are designed to simplify AI programming and make it accessible to a broader audience. Code language models \citep{chen2021evaluating,nijkamp2022codegen,roziere2023code} have enabled AI to automatically generate code that can solve complex tasks, allowing for a deeper integration between programming languages and AI. In the AI agent space,  Flows \citet{Josifoski2023} is conceptual framework that facilitates structured reasoning and collaboration among AI systems through modular, message-based interactions. DSPy\citet{Khattab2023} is a programming model that abstracts language model pipelines as text transformation graphs.\citet{Jojic2023} demonstrate the Iterations by Regimenting Self-Attention  technique, which manipulates the self-attention mechanism in language models to trigger and control iterative behaviors for executing algorithmic tasks. Building on the integration of natural language with programming, the AIOS Compiler \citep{Xu2024} is a system that leverages LLMs to interpret and execute instructions in natural language. AutoGRAMS integrates programming languages and language models by combining graphically represented agent nodes with graphically represented program nodes. computer programs essentially model flowcharts \citep{goldstine1947planning}, where each node in the flow chart executes and instruction and each transition is facilitate by a branch. In AutoGRAMS, the idea of using a flow chart for programming is combined with the idea of using a flowchart to model multi-step language model interactions. The definition of explicit nodes, which can represent conversational states, and transitions between nodes make AutoGRAMS especially well suited for conversational AI applications Transition operations can optionally be governed by multiple choice questions for a language model, and instructions can optionally be governed by prompts to a language model, allowing language models to be deeply embedded in the execution of a program. The visual programming component of an autogram also gives added interpretability, as any autogram can easily be visualized as a graph.

AutoGRAMS is motivated by previous work showing that language models can reason by breaking difficult problems into smaller steps. Chain of thought prompting \citep{Wei2022} showed that prompting LLMs to generate reasoning steps can significantly improve their problem-solving performance of large language models on complex reasoning tasks.  \citet{Kojima2022} further show how a simple prompting strategy, "Let's think step by step," can significantly enhance zero-shot reasoning across various complex tasks.  Tree of Thoughts  \citep{yao2024tree} and Graph of Thoughts framework \citep{besta2024graph} models allow for more complex dependencies of thought units.  Demonstrate-Search-Predict \citep{Khattab2022} incorporates sophisticated interaction pipelines between language models and retrieval models. ReAct \citep{Yao2022} integrates reasoning and action generation in language models. AutoGRAMS gives the flexibility for the designer, which could itself be an autogram, to design specific reasoning steps using thought-type nodes to help the language model solve complex programs. A graph of thought type nodes in combination with multiple choice transitions allows an autogram to represent highly complex thought processes that can be used to solve problems.

Like many previous works, AutoGRAMS enables the ability to incorporate external knowledge tools into an LLM, as well allowing for the LLM to make decisions.  Retrieval-Augmented Generation (RAG) models \citet{Lewis2020}, allow models to access non-parametric memory systems.  Toolformer \citep{schick2024toolformer} is a model that autonomously learns to interact with external APIs . The Automatic Reasoning and Tool-use (ART) framework \citet{Paranjape2023} automates the generation of intermediate reasoning steps and integrates external tool use. AutoGRAMS gives a large amount of flexibility in how external tools are used. Python function nodes can specify the use of an external tool if they call a Python API that applies that tool. The arguments to the Python API can be variables that were set by a language model or another Python API at a previous turn. The node-specific transitions and transition multiple choice questions of AutoGRAMS give additional control over decisions about which tools are selected for use.

AutoGRAMS relates to previous works that use LLMs to engineer prompts that control the execution of tasks.Automatic Prompt Engineer \citep{Zhou2022} automates the generation and optimization of instructions for LLMs and systematically refines prompts through an LLM-driven search and evaluation process. Active-Prompt \citep{Diao2023} uses uncertainty metrics within an active learning framework to optimize chain-of-thought reasoning prompts for LLMs.  Promptbreeder \citep{Fernando2023} uses self-referential self-improvement to evolve and adapt prompts for specific tasks. Likewise-the prompts in AutoGRAMS are also able to result in actions that lead to the modification of an autogram. AutoGRAMS allows any part of an autogram to be modified by the autogram itself, not just the prompts.

Previous approaches have considered meta-learning \citep{hochreiter2001learning,finn2017model}, learning to learn, or meta-programming. Meta-genetic programming \citep{schmidhuber1987evolutionary} relates to our motivation of developing program modifying programs. Unsupervised meta-learning  \citep{metz2018meta} relates to the long term vision of of AutoGRAMS to develop agents that can learn from interactions with their environment in unsupervised ways. \citet{shinn2024reflexion} considers agents that self-reflect on their own mistakes to learn.  Self-Refine \citet{Madaan2023} leverages iterative feedback by enabling an LLM to act as both the generator and refiner.  MetaGPT \citet{Hong2023} is meta-programming framework that enables LLM-based multi-agent collaborations by embedding standardized operating procedures into prompts. While AutoGRAMS itself is a programming language rather than a specific algorithm, its features such the flexibility to fully self-modify, the graphical structure, Turing completeness, and easy integration of language models for executing instructions and determining transitions, make AutoGRAMS for well-suited meta-programming applications. We demonstrate an autogram can be used to design an autogram chatbot from a prompt. AutoGRAMS gives the potential for the development of highly complex meta-agents and self-modifying agents by enabling autograms to design new AutoGRAMS graphs or modify their own graphs.

AutoGRAMS also relates to other approaches for controlling text generation. Previous work has used discriminators \citep{dathathri2019plug,krause2020gedi,yang2021fudge,qin2022cold}, special input tokens \citep{keskar2019ctrl}, prompts \citep{reynolds2021prompt}, instruction following \citep{ouyang2022training,longpre2023flan} or finetuning using a reward signal \citep{ziegler2019fine,rafailov2024direct} to better constrain the outputs of language models for improved controllability and safety. AutoGRAMS allows a complex network of prompts to be defined to control multi-turn or multi-step interactions.

\section{Discussion}

This work presented the AutoGRAMS, a novel framework and high level programming language for controlling multi-step interactions with an LLM. AutoGRAMS gives the designer the ability to create branched multi-step interactions with an LLM in a way that is analogous to how programming languages allow for the design of branched multi-step interactions with a processor. The designer can define these steps of the interaction in advance by writing prompts for a language model at each step. AutoGRAMS also allows language models to exert direct control over transitions in the interaction by enabling the designer to write multiple choice questions that determine which branch to take. This structure enables the designer to use natural language to effectively govern the flow of the of steps that an AutoGRAMS takes. The graphical structure of an autogram allows it to be implemented and visualized in an intuitive way.

AutoGRAMS is especially useful for designing complex conversational flows. AutoGRAMS allows for agents be designed to behave in more predictable ways over longer conversations, giving added control-ability and safety to conversational agents. It accomplishes this by allowing the designer to define a prompt for how to reply to the user for every possible turn in the conversation in advance. The designer has to anticipate the users response, for which The designer can then also design a series of multiple choice questions for a language model that govern which turns will actually be implemented based on that response. Additionally, if the conversational agent needs more complex intermediate reasoning steps or memory, these can be implemented in AutoGRAMS as well. The full state of the autogram can be saved in a serializable object, making it easy to reload the complete state and memory of a complex conversational agent from a database. 

By allowing conditionals and code statements, and variables, AutoGRAMS allows the functionality of a Turing complete programming language to be integrated into the graphs that define the steps of the language model interaction. We demonstrate how Python programs can be mapped to an AutoGRAMS graph of python-type nodes and embedded with other types of nodes. This enables the designer to define data structures to store and access memories, which could be text generated by a language model or variables created in code. These variables can be embedded into prompts, transition questions, referenced in code, or even used to create dynamically executing code or dynamic transitions in the graph. This high level of flexibility makes it possible for AutoGRAMS to be used for very advanced applications of AI agents.

We also demonstrate mechanisms for AutoGRAMS to design other AutoGRAMS. We also demonstrate their ability for self-adaptation and modification. AutoGRAMS allows an agent to be fully self-referential with respect to the nodes that define the agent. We anticipate self-modifying AI agents to be useful for generalizing agents to more situations and allowing agents to learn from experiences and information, and AutoGRAMS provides a useful framework to build such agents.

\subsubsection*{Acknowledgments}
The authors thank Subu Subramaniam for helpful discussions on how autograms could be applied and for helping to test the AutoGRAMS repository.

\bibliography{iclr2021_conference}
\bibliographystyle{iclr2021_conference}
\appendix

\section{Prompt formation}
\label{appx:prompts}

Language models in AutoGRAMS take in a list of input turns (corresponding to user turns or instructions), a list of output turns (corresponding to assistant turns/model outputs), as well as an optional prefix for how the reply should start (for internal models only).

When preparing the prompt to get a new language model output at either a thought node or a chat node, the autogram re-iterates over past turns to get these input and output turns.

For every previously visited node that is within the current scope (depending functions and function types if used), if that node is a thought node or a chat node, it adds a turn to the conversation history that the language model sees. the inputs and outputs are determined as follows:

for each past thought or chat node visited:
\begin{itemize}
    \item if it is a thought node
        \begin{itemize}
        \item  if the thought node came immediately after a user reply, then both the instruction and user reply will appear in the input for that turn

        \item otherwise, only the instruction will appear in the input for that turn
        \end{itemize}

    \item if it is a chat node 
    \begin{itemize}
        \item  if the chat node came immediately after a user reply, then only user reply will appear in the input for that turn. The old instruction is discarded for chat nodes to avoid long prompts. We may later add an option to retain chat instructions in the autogram config

        \item  otherwise,the chat node's instruction will appear in the input for that turn
        \end{itemize}
\end{itemize}

The initial prompt is pre-appended to the first input.

For the last input, corresponding to the end of the prompt, the string is determined by the instruction template. The default instruction template is:

\texttt{<last\_response> Instruction for <agent\_name>: <instruction>}, where \texttt{<last\_response>}, \texttt{<agent\_name>}, and \texttt{<instruction>} are special place holder tokens for the instruction, agent name, and last response. Sometimes the last response will be the empty string if the last user response is already in one of the past inputs (which will happen if there is at least 1 thought node before replying with a chat node).

Lastly, replies have a reply start, which is ``Agent's reply:'' if the default start template and agent name are used. Depending on whether the start type is suffix or prefix, the reply start either appears at the end of the last input, or at the start of the model's reply.

\section{Language models in AutoGRAMS}
\label{appx:language-models}

Every autogram has 3 language models which can share weights or use the same API. The are referred to as the chatbot, classifier, and userbot. The chatbot executes the instruction of thought and chat nodes, the classifier predicts transitions, and the userbot can be used simulate the user in a conversational autogram.

The Chatbot is the language model that generates the output text of chat and thought nodes, corresponding to chain of thought and replies to the user. The chatbot is assumed to be a chat completion style model. As arguments, it takes a list in inputs (treated as user side inputs) and outputs (treated as previous model outputs). Depending on the nodes used, some of the inputs may be instructions or combinations of instructions and user replies, instead of just user replies.  

The classifier model is a language model that predicts the answers to transition questions, and is used to predict transitions. The classifier model only needs to generate a single token, and uses a large positive logit bias on the allowable answer choices (which are always multiple choice A-Z, or yes/no) to ensure the model generates a valid answer. Without a logit bias, the model would likely sometimes generate things like ``The answer is C.'' instead of just ``C'', which would make determining the predicted answer more difficult. Unlike the chatbot, which receives previous inputs and output turns, the conversation history to the classifier is all concatenated to a single turn.

The userbot of an autogram is by default set to be the same object as the chatbot, meaning that it will share the same API or the same weights if running on the local machine. The prompts to the userbot are what distinguish its behavior, since the instructions will typically be along the lines of ``reply as the user and say x''

\section{Interjection Nodes}
\label{appx:interjection}
``Interjection nodes'' are nodes that can be reached after any chat node. They are mostly meant for scenarios where the user does or says something unexpected and it may be necessary to leave the main conversational trajectory. Interjection nodes can be any action type--so a thought node could also be an interjection node. If interjection nodes are used, an additional multiple choice question will be applied after every chat node before applying the transition question. The autogram may override the main transition behavior depending on the answer predicted to that question.

Any node can be made an interjection node by defining a $condition\_interjection$ field when making a node. The default interjection question is ``Which of the following is True?''. The last answer, by default, is ``None of the above.''. The other answer choices correspond to the $condition\_interjection$ set by each of the interjection nodes. These defaults can be changed in the Autogram Config. If any interjection nodes are defined, the classifier will predict the answer to the interjection question after every conversation turn. If the classifier predicts that answer is the default (``none of the above''), the autogram will proceed as normal to ask the language model the transition question. However, if the classifier predicts an answer that corresponds to a $condition\_interjection$, the normal transition behavior will be overridden and the autogram will jump to the interjection node and execute that next. It is important to define the answers for interjection transitions carefully, since if the model incorrectly predicts and interjection transition it can lead to large breaks in the conversational flow.

It would be possible to obtain a similar result to interjection nodes by using additional transition nodes after every chat node. The chat node could ask the interjection question as its transition question and map the answers to the interjection nodes. An additional transition node could be added for when the answer is the default or ``none of the above'', and this transition node could ask the original transition question of the chat node and map to the original transitions. Interjection nodes are essentially a shortcut to achieve this result more conveniently without needing to define an additional transition node after every chat node.

\section{Simulating a user in conversational autograms}
\label{appx:user}

We built the ability to simulate the user's end into conversational autograms. For nodes with multiple transitions, a list of user prompts corresponding to each transition can be provided. The autogram can then be used to simulate entire conversations, using the reply method to simulate the agents end, and the simulate\_user method to reply automatically on the users behalf. Simulate user samples a transition at random and then generates text using the user prompt corresponding to the sampled transition. This simulation ability has several potential usecases. Firstly, it allows a conversational autogram to quickly be tested by generating full conversations automatically that can be inspected to see if the agent is replying in the desired way. Secondly, it can act as a check to be sure the classifier language model used for transitions is predicting the correct output. If the user simulation and classifier both work correctly, then the state sampled by the user simulation should match the transition predicted by the classifier. Lastly, we envision user simulation to be useful for finetuning the underlying language models from human or AI feedback--since it allows for full trajectories to be generated without the need for human input at every step.

\section{Wild card transition visualization}
\label{appx:wildcard}
\begin{figure*}[hbt!]

\begin{center}
\includegraphics[width=.6\textwidth]{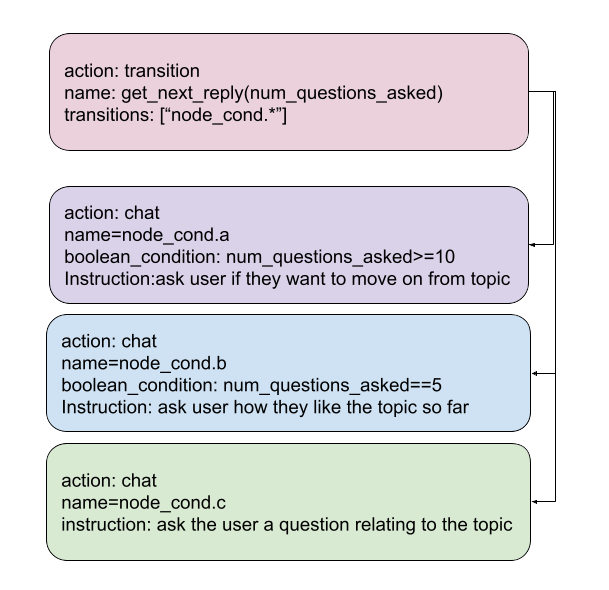}
\end{center}
\caption{Depiction of how wildcard transitions can be used to implement if/else-if/else logic in AutoGRAMS. Any transition ending in ``.*'', such as $node\_cond.*$ in the example above, is interpreted as a transition to one of several possible nodes that have the same prefix but a alphabet character as the ending. In the example above, $node\_cond.*$ can transition to $node\_cond.a$,$node\_cond.b$,and $node\_cond.c$, and the node selected will depend on the $boolean\_condition$ attribute set for these nodes. The boolean condition is evaluated in Python and can include variables visible at the current scope in the autogram. The nodes will be considered in alphabetical order; if the boolean condition for $node\_cond.a$ is true, $node\_cond.a$ will be selected. Otherwise the boolean condition for $node\_cond.b$ will be evaluated, and if its true, $node\_cond.b$ will be selected. If the boolean condition for both $a$ and $b$ are false, then $c$ will be selected. The last node in alphabetical order does not need a boolean condition. This process allows.  }
\label{fig:wildcard}
\end{figure*}

We give the flow chart of nodes to illustrate can wildcard transitions can use if/else-if/else logic to facilitate transitions in figure \ref{fig:wildcard}.

\section{Additional details of the AutoGRAMS graph compiler}
\label{appx:compiler}

The AutoGRAMS compiler translates code with python-like syntax into a representation that the AutoGRAMS interpreter can understand. It allows for Python statements, conditionals, loops, variable assignments, and function calls, to be combined with special $exec\_node()$ statements that create AutoGRAMS nodes at specific points in the program. T AutoGRAMS compiler processes code down to the level of individual nodes and statements, and to form an AutoGRAMS graph around these statements. It uses the Python abstract syntax tree (AST) to parse the code, and recursively traverses the AST until it reaches code that can be represented as a single statement. It then uses this statement to define a new AutoGRAMS node. If the statement is a Python statement, the compiler will define a python-type node, whereas if it is a special $exec\_node()$ statement it will be a node type defined by the arguments given to $exec\_node()$

The high level steps implemented by the AutoGRAMS compiler are:

\begin{enumerate}
    
\item Convert code to Python AST representation
\item Extract the name of every function definition in the code and the subtree of the AST associated with that function
\item Recursively traverse the subtree of the AST for each function to get the AutoGRAMS graph for that function
\item Recursively traverse the AST for code that is outside of a function to get the AutoGRAMS graph of the outer scope/main program
\end{enumerate}

Any nodes that aren't explicitly named by \texttt{exec\_node} statements need to be named automatically by the AutoGRAMS compiler. Node names need to be unique, so the compiler assigns names with prefixes based on functions and statement bodies that they reside in, and using a numbering system that increments for each statement type and node within each prefix. All automatically named names start with the'\_' character to reduce the risk of conflicts between nodes named by the designer and nodes named automatically by the compiler.

As we define the operations performed during AST traversal, we will use ``AST node'' to refer to a node in the AST tree and ``AutoGRAMS node'' to refer to e anode in the AutoGRAMS graph. The AutoGRAMS compile recursively traverses the AST tree, starting at the root AST node, and recursively traversing the subtree of the children AST nodes.

When encountering an AST node representing a conditional (if/else-if/else), the compiler creates a new independent chain object for each branch to recursively processes each branch of the conditional. Each chain returns with the AutoGRAMS nodes needed to implement that branch. The subgraphs in each branch are then combined by defining new AutoGRAMS transition nodes at the root of the conditional, and at the start of each branch of the conditional. The AutoGRAMS node at the root of the conditional is given an automatically defined wildcard style transition, and the AutoGRAMS nodes at the start of each branch are given names that match the name defined in the wildcard style transition, and a alphabet character suffix corresponding to their order in the branch logic. These AutoGRAMS nodes are also defined with boolean conditions that match the boolean condition defined in the Python conditional statement. This combination of a wild card transition, node names, and boolean conditions allows them to implement the exact transition logic of the if/else statement defined in Python.

When encountering an AST node corresponding to a while loop, the compiler first collects the condition or the while loop. It then creates a new independent chain object that recursively processes the body of the while loop to obtain the AutoGRAMS nodes in the body.The compiler can then follow the process to design conditionals described above, and can connect the final AutoGRAMS node in the body of the while loop with root AutoGRAMS node of the conditional, allowing the loop of the graph to be repeatedly traversed until the exit condition is met.

When encountering an AST node corresponding to a while loop, the compiler can execute a similar process to a while loop, while also creating an AutoGrams node to make node before the loop to define an iterator variable, as well as AutoGrams nodes in the forloop that extracts elements from the iterable object defined in the forloop, and increment the iterator variable. The forloop is given an exit condition that causes it to exit the graph when the iterator variable is equal to the length of the iterable object.

When encountering AST nodes corresponding to single Python statements, the autograms compiler adds a python-type $python\_function$ node to cause that statement to be executed at that point in the graph. If the statement contains a varaible assignment, this is also added to the instruction off the corresponding AutoGRAMS node.

When encountering $exec\_node$ statements, the autograms compiler defines a new node with the arguments given in the statement. If the exec node statement contains a pythnon style variable assignment, this assignment is added to the node's instruction string. If the $exec\_node$ statement contains no defined transitions, it is assumed that it will transition to the next node encountered in the code. It is also possible to connect an $exec\_node$ statement with nodes defined by other $exec\_node$ statements, however jumping into a loop or function from an external point will lead to errors.

\section{The reply algorithm}
\label{appx:reply-algo}

We present high-level pseudo-code to show the operations performed by the AutoGRAMS reply method to get a conversational response and updated memory from an autogram.

\begin{algorithm}

\caption{autogram.reply() method. Some arguments to functions are omitted for simplicity.}
\begin{algorithmic}[1]
\Function{$reply$}{$user\_reply,memory\_object$,$new\_node\_id$}
\State $response\_to\_user \gets {False}$
\While{$ !({response\_to\_user})$}
\If{$new\_node\_id \,\,is\,\, None$}
\\
\Comment{get output of previously executed node}
\State $variable\_output \gets node.get\_variable\_output(memory\_object)$
\\
\Comment{assign variables in memory}
\State $memory\_object.assign\_variables(variable\_output)$
\\
\Comment{apply transition function}
\State $new\_node\_id \gets node.apply\_transition(user\_reply,memory\_object)$
\\
\Comment{process transition string returned by transition function}
\State $new\_node\_id \gets process\_node\_id(new\_node\_id,memory\_object)$
\EndIf
\Comment{select new node}
\State $node \gets self.nodes[new\_node\_id]$
\\
\Comment{executes node's function, which will call a language model for chat or thought style nodes}
\State $response,user\_reply,response\_to\_user \gets$

\Statex \hspace{3.5em} $node.apply\_instruction(user\_reply,memory\_object,self.chatbot)$

\EndWhile
\State $\textbf{return}\,\,response,memory\_object$

\EndFunction
\end{algorithmic}
\end{algorithm}

\clearpage
When only using chat style nodes with only standard transitions, the algorithm for replying to a user becomes simplified, allowing it to be more fully explained in a single block of pseudo-code. This pseudo-code assumes that it is not the first step in the conversation (if the model goes first, it will reply from a predetermined node).

\begin{algorithm}
\caption{A simplified pseudo-code for reply algorithm with only chat-style nodes}
\begin{algorithmic}[1]
\Function{$reply$}{$user\_reply$,$conv\_history$,$old\_node\_id$}

\State $node \gets nodes[old\_node\_id]$
\State $conv\_hisory.add\_user\_reply(user\_reply)$

\If{$len(node.transition\_choices)>1$}

\State $transition\_question \gets  node.transition\_question$
\State $transition\_choices \gets node.transition\_choices$

\Comment{Format as a multiple choice question of the form ``\textbf{Question}  A. \textbf{choice 1} B. \textbf{choice 2} C. \textbf{choice 3}'' etc.}
\State $mc\_question \gets form\_mc\_prompt(transition\_question,transition\_choices)$

\State $classifier\_prompt \gets form\_classifier\_prompt(conv\_history,mc\_question)$

\Comment{classifier is a language model that predicts probability distribution over a single next token. It will predict A, B, C etc. depending on number of transition choices.}
 \State $next\_token\_probs \gets classifier\_lm.predict\_next\_token (classifier\_prompt)$

\Comment{get token ids corresponding to capital letters that model language will predict for answer choice}
\State $mc\_indices \gets ABCD\_TOKENS[:len(transition\_choices)]$

\Comment{only select probabilities corresponding to allowable predictions}
\State $abcd\_probs \gets next\_token\_probs[mc\_indices]$

\Comment{get the answer choice the language model thinks is most likely}
\State $answer\_pred \gets argmax(abcd\_probs)$

\Comment{select the next node to visit}
\State $new\_node\_id \gets node.transitions[answer\_pred]$
\Else
\State $new\_node\_id \gets node.transitions[0]$
\State $node \gets nodes[new\_node\_id]$

\EndIf

 \State $chatbot\_prompt \gets form\_chatbot\_prompt(conv\_history,node.instruction)$
 \\
\Comment{apply chatbot to generate response}
\State $agent\_reply \gets chatbot\_lm(chatbot\_prompt)$

 \State $conv\_history.add\_agent\_reply(agent\_reply)$
\State $response \gets node.apply\_instruction(user\_reply,self.chatbot)$

\State $\textbf{return}\,\,response,conv\_history,new\_node\_id$

\EndFunction
\end{algorithmic}
\end{algorithm}

\section{A simple self-referential autogram}
\label{appx:self-ref}

We demonstrate a very simple proof-of-concept of a self-referential autogram below. This autogram creates a new node at every conversation turn to handle the user's reply. the autogram was implemented in AutoGRAMS compiled from Python and automatically compiled into AutoGRAMS nodes. The fields for the nodes that make up the autogram are given below.

The first node sets a variable to hold an instruction we will use later.

\begin{verbatim}
name: "_node1"
action: "python_function"
transitions: ['intro']
instruction: "meta_inst = \"We need to decide how to 
   deal with the last user reply. 
   Rather than replying directly, 
   write an instruction for another language model 
   to reply.The instruction could be 
   along the lines of 
   'respond to the user and tell them xyz'\""
\end{verbatim}

the next node initiates a conversation turn with the user with the exact text "Hi there, What can i help you with?".
\begin{verbatim}
name: "intro"
action: "chat_exact"
instruction: "Hi there, What can i help you with?"
transitions: ['dynamic_node']

\end{verbatim}

The next node calls a function that will do the majority of the heavy lifting. The function \texttt{add\_new\_node()} will add a new node to the autogram. \texttt{add\_new\_node()} also returns the string of the next node, which will be saved in the variable \texttt{next\_node}. Since transitions are executed after instructions, the transition \texttt{\$next\_node} will cause the autogram to transition to the node defined by the $add\_new\_node$ function.
\begin{verbatim}
name: "dynamic_node"
action: "function"
transitions: ['$next_node']
instruction: "next_node=add_new_node(meta_inst)"
\end{verbatim}

The next node defines a callable node that starts the \text{add\_new\_node} function, with the name field defining the function name and arguments.  
\begin{verbatim}
name: add_new_node(meta_inst)
action: transition
transitions: ['_add_new_node_node2']
\end{verbatim}

The next node is a thought node that uses the language model to generate an instruction for the new node are are designing. It conditions on the variable \texttt{meta\_inst} as a prompt. The resulting generated text is saved in a new variable called \texttt{new\_inst}.
\begin{verbatim}
name: _add_new_node_node2
action: thought
instruction: new_inst=$meta_inst
transitions: ['_add_new_node_node3']


\end{verbatim}

The next node is a thought node that uses the language model to name the new node being created. It conditions on the new instruction generated at the previous node, which is embeded in the prompt. The result is saved in a  variable called \texttt{new\_node}

\begin{verbatim}
name: _add_new_node_node3
action: thought
instruction: new_name=We need to come up with a short 
     name for a node that executes this instruction: $new_inst
     The name must be all lower case letters and underscore ('_'). 
     Reply with the name only with no spaces
transitions: ['_add_new_node_whileloop1_start']
\end{verbatim}

the next node is an automatically generated node to facilitate the start of a while loop by applying a wildcard transition to check the exit conditions, which will be set in the \texttt{boolean\_condition} attributes of 2 later nodes
\begin{verbatim}
name: _add_new_node_whileloop1_start
action: transition
transitions: ['_add_new_node_whileloop1.*']
\end{verbatim}

The first condition referenced by the wildcard transition. This will enter the loop if the node name we have defined not valid.

\begin{verbatim}
name: _add_new_node_whileloop1.a
action: transition
transitions: ['_add_new_node_whileloop1_node1']
boolean_condition: not meta_utils.check_node_name(new_name)
\end{verbatim}

This node regenerates node name if node name from before was invalid.

\begin{verbatim}
name: _add_new_node_whileloop1_node1
action: thought
instruction: new_name=We need to come up with a short 
     name for a node that executes this instruction: $new_inst
     The name must be all lower case letters and underscore ('_'). 
     The last name was invalid, please be sure to reply with 
     the name only and obey the rules about characters.
transitions: ['_add_new_node_whileloop1.*']
\end{verbatim}

this node is to exit the while loop when the boolean condition set in \texttt{\_add\_new\_node\_whileloop1.a} is False.

\begin{verbatim}
name: _add_new_node_whileloop1.b
action: transition
transitions: ['_add_new_node_node8']
\end{verbatim}

This node adds the node that we just defined the arguments for to the autogram interpreter object that is executing the graph, which is referenced using \text{self}. We set the action of this node to \texttt{chat}, since we will use it to give conversational replies. Notice that it transitions back to \texttt{dynamic\_node} which we defined earlier. This ensures that the autogram will continue--after the generated node is executed, the autogram will transition back to \texttt{dynamic\_node} and design another node to handle the new conversation turn. 
\begin{verbatim}
    
name: _add_new_node_node8
action: python_function
instruction: self.add_node(action='chat',
     instruction=new_inst, 
     name=new_name, 
     transitions=['dynamic_node'])
transitions: ['_add_new_node_node9']

\end{verbatim}

Lastly, a node with a return transition returns the \texttt{new\_name} variable. This returns the graph to \texttt{dynamic\_node}, which called the function. 

\begin{verbatim}
name: _add_new_node_node9
action: transition
transitions: ['return new_name']
\end{verbatim}

After the function returns, \texttt{dynamic\_node} will then transition to the new node we just defined (since it's name is saved in the \texttt{new\_name} variable and  \texttt{dynamic\_node} uses \texttt{\$new\_name} as its only transition. The new node will execute to give a conversational reply, and then transition back to \texttt{dynamic\_node}, which will call the function again to design another node.

\begin{figure*}[hbt!]

\begin{center}
\includegraphics[width=1\textwidth]{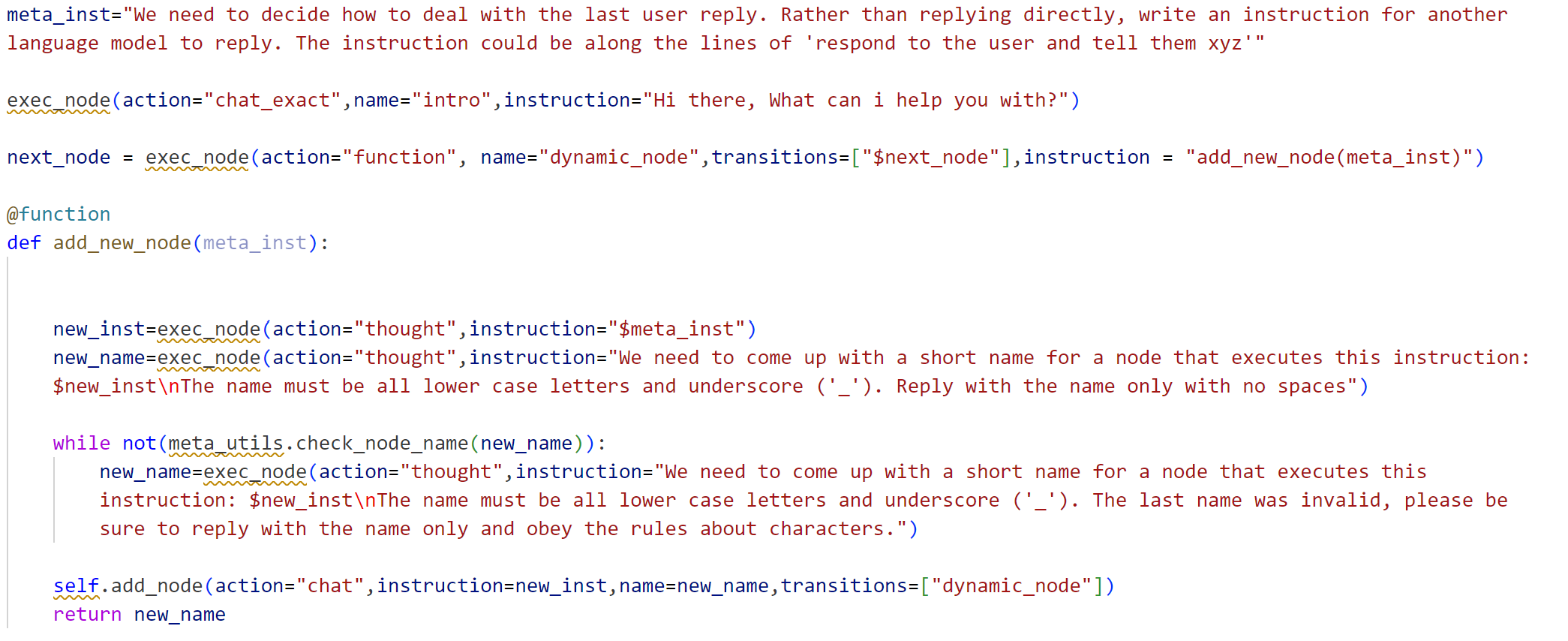}
\end{center}
\caption{AutoGRAMS compiled from Python code for self-referential autogram described in Appendix \ref{appx:self-ref}. The autogram implements a function that designs a new node, and modifies its own representation with the self.add\_node() method. The name of the new node is returned and set as the transition. The newly designing node gives a conversational reply, and transitions back to the node-designing node, allowing the autogram to design a new node at each conversation turn. }
\label{fig:self-ref}
\end{figure*}

\end{document}